\documentclass[times,referee,twocolumn,final,authoryear]{elsarticle}

\usepackage{ycviu}
\usepackage{framed,multirow}

\usepackage{amssymb}
\usepackage{latexsym}

\usepackage{url}
\usepackage{xcolor}
\definecolor{newcolor}{rgb}{.8,.349,.1}

\usepackage{bm}
\usepackage{amsmath}
\usepackage{mathtools}
\usepackage{pgfplots}
\pgfplotsset{compat=1.14}
\usepackage{silence}
\usepackage{subcaption}
\usepackage{booktabs}
\usepackage{microtype}
\usepackage{hyperref}
\usepackage[capitalise]{cleveref}

\usepackage{authorbiography}

\DeclareMathOperator*{\argmin}{arg\,min}
\DeclarePairedDelimiterX{\infdivx}[2]{(}{)}{%
    #1\;\delimsize\|\;#2%
}
\DeclarePairedDelimiter{\abs}{\lvert}{\rvert}
\newcommand{\kldiv}{D_\text{KL}\infdivx}
\DeclarePairedDelimiter\norm{\lVert}{\rVert}%

\makeatletter
\let\oldabs\abs
\def\abs{\@ifstar{\oldabs}{\oldabs*}}
\let\oldnorm\norm
\def\norm{\@ifstar{\oldnorm}{\oldnorm*}}
\makeatother

\newcommand{\approximately}{\raise.17ex\hbox{$\scriptstyle\sim$}}

\usetikzlibrary{math}
\tikzmath
{
    function symlog(\x,\a){
        \yLarge = ((\x>\a) - (\x<-\a)) * (ln(max(abs(\x/\a),1)) + 1);
        \ySmall = (\x >= -\a) * (\x <= \a) * \x / \a ;
        return \yLarge + \ySmall ;
    };
    function symexp(\y,\a){
        \xLarge = ((\y>1) - (\y<-1)) * \a * exp(abs(\y) - 1) ;
        \xSmall = (\y>=-1) * (\y<=1) * \a * \y ;
        return \xLarge + \xSmall ;
    };
}

\journal{Computer Vision and Image Understanding}

\begin{document}

\begin{frontmatter}

\title{Generalizing semi-supervised generative adversarial\\networks to regression using feature contrasting}

\author[1,2]{Greg \snm{Olmschenk}}\corref{cor1}
\cortext[cor1]{Corresponding author: Tel.: +1-651-366-1814;}
\ead{golmschenk@gradcenter.cuny.edu}
\author[1,2]{Zhigang \snm{Zhu}}
\author[3]{Hao \snm{Tang}}

\address[1]{The City College, The City University of New York, 160 Convent Ave, New York, NY 10031, USA}
\address[2]{The Graduate Center, The City University of New York, 365 5th Ave, New York, NY 10016, USA}
\address[3]{Borough of Manhattan Community College, The City University of New York, 199 Chambers St, New York, NY 10007, USA}

\received{1 May 2013}
\finalform{10 May 2013}
\accepted{13 May 2013}
\availableonline{15 May 2013}
\communicated{S. Sarkar}

\begin{abstract}
In this work, we generalize semi-supervised generative adversarial networks (GANs) from classification problems to regression problems. In the last few years, the importance of improving the training of neural networks using semi-supervised training has been demonstrated for classification problems. We present a novel loss function, called feature contrasting, resulting in a discriminator which can distinguish between fake and real data based on feature statistics. This method avoids potential biases and limitations of alternative approaches. The generalization of semi-supervised GANs to the regime of regression problems of opens their use to countless applications as well as providing an avenue for a deeper understanding of how GANs function. We first demonstrate the capabilities of semi-supervised regression GANs on a toy dataset which allows for a detailed understanding of how they operate in various circumstances. This toy dataset is used to provide a theoretical basis of the semi-supervised regression GAN. We then apply the semi-supervised regression GANs to a number of real-world computer vision applications: age estimation, driving steering angle prediction, and crowd counting from single images. We perform extensive tests of what accuracy can be achieved with significantly reduced annotated data. Through the combination of the theoretical example and real-world scenarios, we demonstrate how semi-supervised GANs can be generalized to regression problems.
\end{abstract}


\end{frontmatter}

\section{Introduction}

Deep learning~\citep{lecun2015deep}, particularly deep neural networks (DNNs), has become the dominant focus in many areas of computer science in recent years. This is especially true in computer vision, where the advent of convolutional neural networks (CNNs)~\citep{lecun1999object} has led to algorithms which can outperform humans in many vision tasks~\citep{dodge2017study}. Within the field of deep learning, generative models have become popular for generating data that simulates real datasets. A generative model is one which learns how to produce samples from a data distribution. In the case of computer vision, this is often a neural network which learns how to generate images, possibly with specified characteristics. Generative models are particularly interesting because for such a model to generate new examples of data from a distribution, the model must be able to distinguish data which belongs to the distribution and that which does not. In a sense, this distinguishing ability shows that the network "understands" a data distribution. Arguably the most powerful type of generative model is the generative adversarial network (GAN)~\citep{goodfellow2016nips, goodfellow2014generative}. GANs have been shown to be capable of producing fake data that appears to be real to human evaluators. For example, GANs can generate fake images of real world objects which a human evaluator can not distinguish from true images~\citep{elsayed2018adversarial}. Beyond this, GANs have been shown to produce better results in discriminative tasks using relatively small amounts of data~\citep{salimans2016improved}, where equivalent DNNs/CNNs would require significantly more training data to accomplish the same level of accuracy. As one of the greatest obstacles in deep learning is acquiring the large amount of labeled data to train such models, the ability to train these powerful models with much less data is of immense importance.

While GANs have already shown significant potential in semi-supervised training, they have only been used for a limited number of cases. In particular, they have almost exclusively been used for classification problems thus far. In this work, we propose generalizing semi-supervised GANs to regression problems. Though this may initially seem to be a trivial expansion, the nature of a GAN's optimization goals makes the shift from classification to regression problems difficult. Specifically, the two parts of a GAN can be seen as playing a minimax game. The discriminating portion of the GAN must have the objective of labeling the fake data from generating portion as fake. In a classification semi-supervised GANs, an additional "fake" class is added to the possible list of classes. However, in regression, where the data is labeled with real valued numbers, deciding what constitutes a "fake" labeling is not straight forward.

\subsection{Contributions}
In this work, we will present the following contributions:
\begin{enumerate}
    \item A new algorithm with a novel loss function, feature contrasting, which allows semi-supervised GANs to be applied to regression problems, the Semi-supervised Regression GAN (SR-GAN).
    \item A set of optimization rules which allows for stable, consistent training when using the SR-GAN, including experiments demonstrating the importance of these rules.
    \item Systematic experiments using the SR-GAN on the real world applications of age estimation, driving steering angle prediction, and crowd counting from single images showing the benefits of SR-GANs over existing approaches.
\end{enumerate}

The most important contribution is the introduction of the generalized semi-supervised regression GAN (SR-GAN) formulation using feature contrasting. Nevertheless, while the theoretical solution for applying semi-supervised GANs to regression is provided in the first contribution, there are several factors that need to be addressed for this approach to work in practice. Chiefly is the stability of training the two competing networks in an SR-GAN. This is addressed by designing loss functions for the SR-GAN whose gradients are well-behaved (neither vanishing nor exploding) in as many situations as possible, and preventing cyclical training between the generator and discriminator by applying penalties and limitations in the training behavior.

We provide a number of real world applications where SR-GANs are shown to improve the results over traditional CNNs and other competing models. Specifically we will use the SR-GAN to predict the age of an individual, estimate the angle a steering wheel should be turned to given an image of the upcoming road segment, and count the size of a crowd from a single image. The age estimation and steering angle datasets provides relatively simple applications on which the SR-GAN can be used to reduce the data requirements in a real world situation, while still being challenging and general enough to merit attention. The crowd counting application provides a more complex scenario with a wide variety of conditions to show the method in more difficult circumstances.

\subsection{Outline}
The remainder of the paper is laid out as follows. The work which our method builds off of as a starting point and other related works are examined in \cref{sec:Background and Related Work}. \cref{sec:Theory and Design} explains our methods and experimental setup. \cref{sec:Results} displays the experimental results and discusses the findings. Finally, we conclude in \cref{sec:Conclusion}.
\section{Background and Related Work}
\label{sec:Background and Related Work}

\subsection{The Value of Regression Problems}
Regression problems encompass a large pool of applications which cannot be solved--or would be poorly solved--by framing them as classification problems. The SR-GAN as we define it here can be generalized to any such regression problem. Some examples include crowd counting estimation~\citep{zhang2015cross}, weather prediction models~\citep{xingjian2015convolutional}, stock index evaluation~\citep{ding2015deep}, object distance estimation~\citep{eigen2014depth}, age estimation~\citep{niu2016ordinal}, data hole filling~\citep{pathak2016context}, curve coefficient estimation, ecological biomass prediction~\citep{ali2015review}, traffic flow density prediction~\citep{lv2015traffic}, orbital mechanics predictions~\citep{hartikainen2012state}, electrical grid load prediction~\citep{marino2016building}, stellar spectral analysis~\citep{fabbro2017application}, network data load prediction~\citep{oliveira2016computer}, object orientation estimation~\citep{schwarz2015rgb}, species population prediction~\citep{bland2015predicting}, ocean current prediction~\citep{liu2005patterns}, and countless others. While it is possible to frame each of these problems in terms of classification, in practice, this presents several significant problems. For example, the developer must decide on an arbitrary number of classes for the application. However, more importantly, such a naive classification approach results in each incorrect prediction being considered equally as erroneous. In regression applications, the true label lies somewhere on a continuous scale, and the closer of two predictions should always be considered better than the farther, even if both are inaccurate. If the prediction of a real number from 0 to 10 was split into 10 discrete classes, a prediction of 8 should be considered better than a prediction of 2 for a true label of 10. Yet, a naive classification network produces the same loss for each. Depending on the accuracy required by the application, this approach may be acceptable, but these problems are more naturally framed as regression problems.

\subsection{Generative Adversarial Networks}

A Generative Adversarial Network (GAN) consists of two neural networks which compete against one another. One of the networks generates fake data; hence we will call it the generator. The other network attempts to distinguish between real data and the fake generated data; consequently, this network is called the discriminator. Both networks are trained together, each continually working to outperform the other and adapting in accordance to the other.

\begin{figure}
    \centering
    \includegraphics[width=0.65\columnwidth]{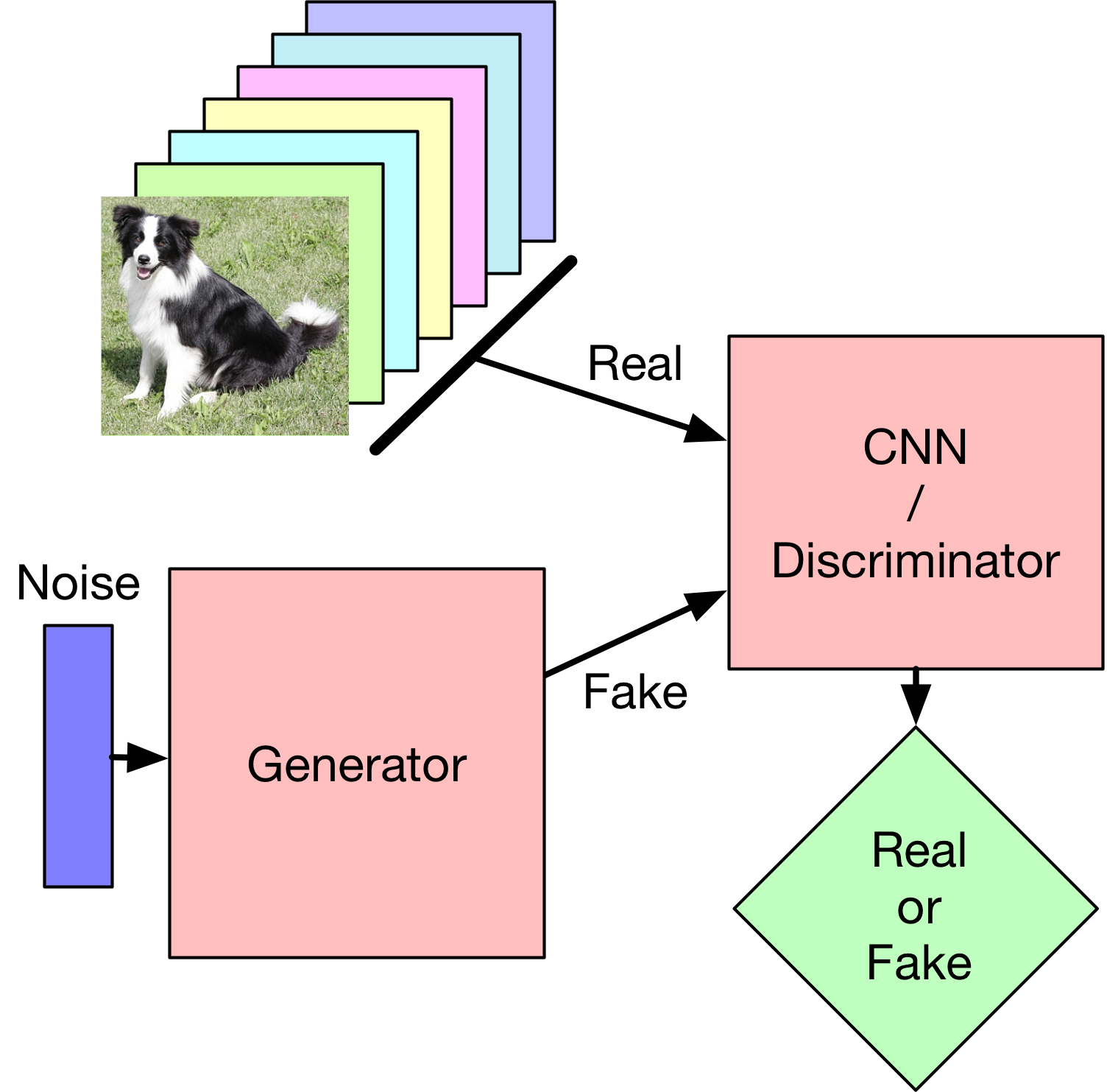}
    \caption{The structure of a basic GAN. Real and fake images are fed to a discriminator network, which tries to determine whether the images are real or fake. The fake images are produced by a generator network.}
    \label{fig:GAN Diagram}
\end{figure}

Though GANs are now fairly common, to provide the groundwork for our SR-GAN, it is worth defining the details of a GAN from the viewpoint of probability distributions. Although these methods work for any prediction application, to give a concrete understanding, these explanations are given in terms of computer vision problems, specifically where the datasets consist of images. This means an example of real data (and thus the input of the discriminator) is an image, and the output of the generator is also an image. The structure of a GAN can be seen in \cref{fig:GAN Diagram}.

The generator network takes random noise as input (usually sampled from a normal distribution) and outputs the fake image data. The discriminator takes as input images and outputs a binary classification of either fake or real data. Images can be represented by a vector, with each element representing the value of a pixel in the image\footnote{One element per pixel is in the case of grayscale images. For RGB images, there will be three elements in the vector for each pixel, one for each color channel of the pixel.}. In any image, each element of this vector has a value within a certain range representing the intensity of that pixel. For this explanation, we will state the minimum element value (pixel value) as being 0, and the maximum as being 1. Of course, this vector can be represented as a point in $N$ dimensional space, where $N$ is the number of elements in the vector. The possible positions of an image's point are restricted to the $N$ dimensional hypercube with a side length of 1. Here, it is important to note that real-world images are not equally spread throughout this cube. That is, most points in the cube correspond to images that would look like random noise to a human. Images from the real world usually have properties like local consistency in both texture and color, logical relative positioning of shapes, etc. Real world images lie on a manifold within the cube~\citep{fefferman2016testing}. Subsets of real-world images, such as the set of all images containing a dog, lie on yet a smaller manifold. This manifold represents a probability distribution of the real world images. We can view the real world as a data generating probability distribution, with each position on the manifold having a certain probability based on how likely that image is to exist in the real world.

The goal of the generator is then to produce images which match the probability distribution of the manifold as closely as possible. Input to the generator is a point sampled from the probability distribution of (multidimensional) random normal noise, and the output is a point in the hypercube--an image.  The generator is then a function which transforms a normal distribution into an image data distribution. Formally,
\begin{equation}
p_{fake}(\bm{x}) = G(\mathcal{N})
\end{equation}
where $G$ represents the generator function, $\bm{x}$ is a random variable representing an image, $\mathcal{N}$ is the normal distribution, and $p_{G}(\bm{x})$ is the probability distribution of the images generated by the generator. The desired goal of the generator is to minimize the difference between the generated distribution and the true data distribution. One of the most common metrics to minimize this difference is the Kullback-Leibler (KL) divergence between the generator distribution and the true data distribution using maximum likelihood estimation. This is done by finding the parameters of the generator, $\bm{\theta}$, which produce the smallest divergence,
\begin{equation}
\bm{\theta}^* = \argmin_{\bm{\theta}} \kldiv{p_{data}(\bm{x})}{p_{G}(\bm{x}; \bm{\theta})} \text{.}
\end{equation}
To find this set of parameters, each of the discriminator and the generator  works toward minimizing a loss function. For the discriminator, the loss function is given by
\begin{equation}
L_{D} = -\mathbb{E}_{\bm{x} \sim p_{data}(\bm{x})}[\text{log}D(\bm{x})] - \mathbb{E}_{\bm{x} \sim p_{fake}(\bm{x})}[\text{log}(1-D(\bm{x}))]
\end{equation}
and the generator's loss function is given by
\begin{equation}
L_{fake} = -\mathbb{E}_{\bm{x} \sim p_{fake}(\bm{x})}[\text{log}(D(\bm{x}))] \text{.}
\end{equation}
In the case of image data, this approach has led to generative models which can produce realistic looking images reliably~\citep{radford2015unsupervised}.

\subsection{Semi-Supervised GANs for Classification}
\begin{figure}
    \centering
    \includegraphics[width=0.65\columnwidth]{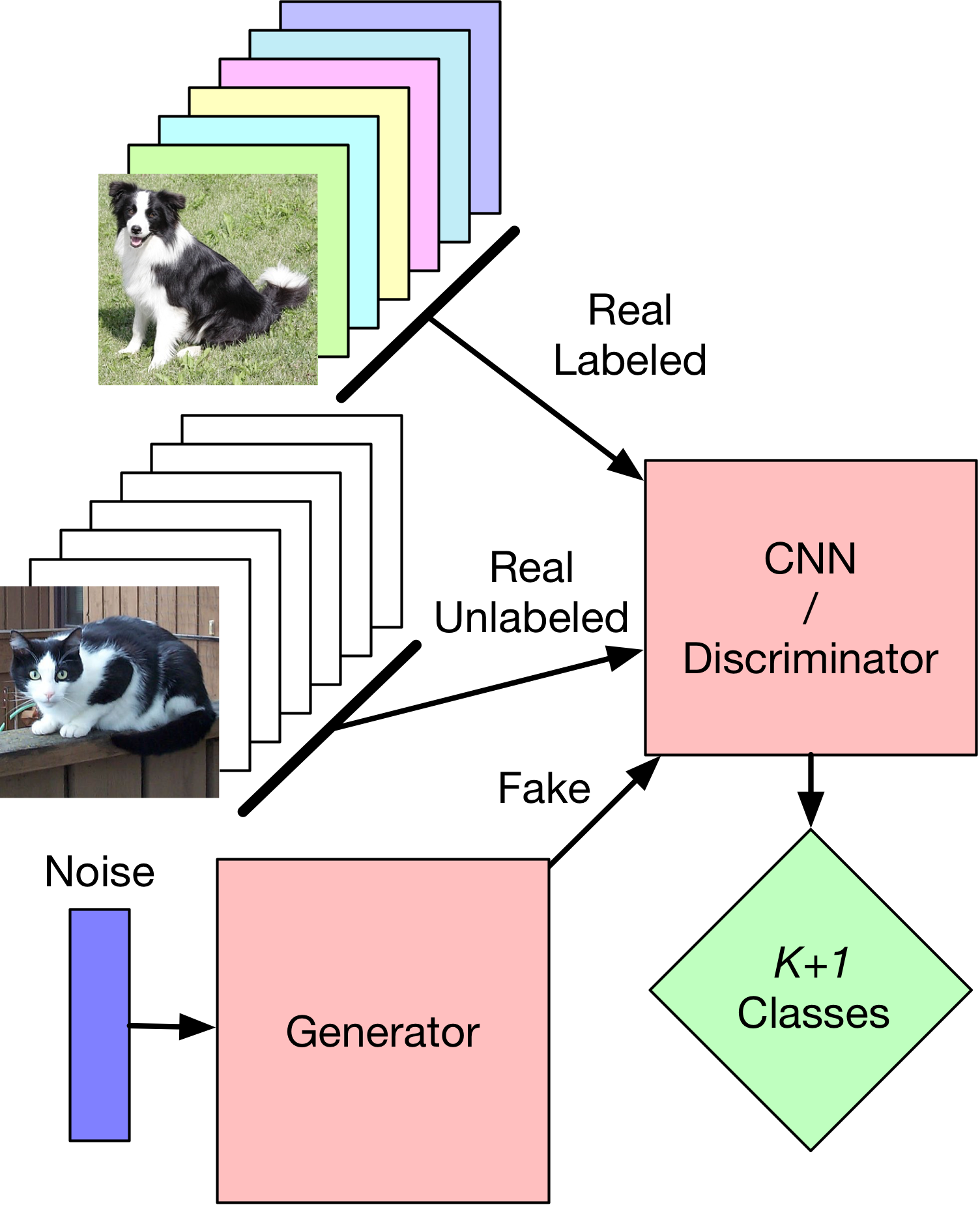}
    \caption{The structure of a semi-supervised GAN. Both labeled and unlabeled real images, as well as fake images, are fed to a discriminator network, which tries to determine which class each image belongs to (K real classes and one fake class). The discriminator wishes to label images from the generator as belonging to a special "fake" class.}
    \label{fig:S-GAN Diagram}
\end{figure}

In this section, we will explain a subset of GANs which are used to improve the training of ordinary networks for discrimination and prediction tasks. In this case, both a labeled and an unlabeled dataset is used, and in addition to distinguishing between real and fake, the discriminator also tries to label a real input data sample  into one of the given classes. The primary goal of this type of GAN is to allow the discriminator's prediction task to be trained with relatively small amounts of labeled data using unlabeled data to provide the network with additional information. As unlabeled data is usually much easier to obtain than labeled data, this provides a powerful means to reduce the requirements of training neural networks. This semi-supervised GAN structure can be seen in \cref{fig:S-GAN Diagram}.

Where in a simple GAN the discriminator would be passed true examples and fake examples, in the semi-supervised GAN the discriminator is given true labeled examples, true unlabeled examples, and fake examples. We can better understand why this is useful by considering the case of image classification. In this case, the discriminator is being trained to predict the correct class of a true image, which can be one of the $K$ classes that exist in the dataset. The discriminator is given the additional goal of attempting to label any fake images with a $K+1$th class, which only exists to label fake data (i.e., does not exist in the true label dataset). For the case of unlabeled, all we know is that it must belong to one of the first $K$ classes, as the $K+1$th class does not exist in the real data. The discriminator is then punished for labeling true unlabeled data as the $K+1$th class. This is useful because the discriminator cannot simply overfit to the labeled data, as it still has to accommodate for the unlabeled data. At the same time, the fake data prevents the discriminator from allowing simple features to be the deciding factor, as the generator is able to produce such simple features.

To understand what is happening in this semi-supervised learning more intuitively, we can imagine the extreme case of an ideal discriminator and generator. The generator would have to have learned to produce data which exactly matches the true data distribution. For this to happen, the discriminator must have forced the generator to learn this (as the generator's training is entirely dictated by backpropagation from the discriminator), meaning the discriminator too "knows" exactly the data distribution. If there were any difference between the true and generated image distributions, the discriminator could use this to distinguish between real and fake, and then the generator could still be trained further toward producing a match of the true distribution.

Viewing this from the perspective of the manifold in data space again, there are few labeled data points and many unlabeled data points which must lie on the manifold. The manifold has different regions (or even separate manifolds) for each class, but even the unlabeled data has to lie somewhere on the manifold. As the discriminator trains, it learns how to segment the data points into categories. To do this, it creates a mapping from a predictive manifold to a class, with the training warping the manifold to contain each of the data points for that class. At the same time, the generator prevents the manifold from warping too severely to reach data points in arbitrary ways. Intuitively, this is because severely warping the manifold to reach true data points can result in the manifold stretching into the area which does not represent true images. The generator acts a pressure on the manifold to reduce this. By generating images near the manifold, the generator forces the discriminator's manifold not to wander into areas that don't contain real images. In this sense, the generator is a form of regularization for the discriminator, but one which is based on real-world data.

As originally formulated by \citet{salimans2016improved}, the discriminator loss function is then defined by
\begin{equation}
L_{D} = L_{supervised} + L_{unsupervised}
\label{eq:classification_discriminator_loss}
\end{equation}
\begin{equation}
\begin{split}
L&_{supervised} = \\
& -\mathbb{E}_{\bm{x},y \sim p_{labeled}(\bm{x},y)}\text{log}[p_{model}(y \mid \bm{x}, y < K+1)]
\end{split}
\label{eq:classification_discriminator_supervised_loss}
\end{equation}
\begin{equation}
\begin{split}
L&_{unsupervised} = \\
& -\mathbb{E}_{\bm{x} \sim p_{unlabeled}(\bm{x})}\text{log}[1 - p_{model}(y = K+1 \mid \bm{x})] \\
& -\mathbb{E}_{\bm{x} \sim p_{fake}}\text{log}[p_{model}(y = K+1 \mid \bm{x})]
\text{.}
\end{split}
\label{eq:classification_discriminator_unsupervised_loss}
\end{equation}

As for the generator, the first option for a loss function is the straight forward one which aims to have the discriminator label the fake images as from real classes. Specifically,
\begin{equation}
L_{G} = -\mathbb{E}_{\bm{x} \sim p_{fake}}\text{log}[p_{model}(y < K+1 \mid \bm{x})] \text{.}
\end{equation}
However, \citet{salimans2016improved} found better results by trying to have the output activations of an intermediate layer of the discriminator have similar statistics in both the fake and real image cases. That is, the generator should try to make its images produce similar features in an intermediate layer as is produced when true images are input. This can be intuitively understood as making the statistics of the image be the same in both the fake and real cases, specifically, the feature statistics that are used in deciding a classification. The simplest and most useful statistic to try to match is the expected value for each feature. Formally put, if we denote $f(\bm{x})$ as the features output by an intermediate layer in the discriminator, then the loss function for the generator becomes
\begin{equation}
L_{G} = \norm{
    \mathbb{E}_{\bm{x} \sim p_{real}} f(\bm{x}) -
    \mathbb{E}_{\bm{x} \sim p_{fake}} f(\bm{x})
}^2_2
\text{.}
\end{equation}

Since their development, semi-supervised GANs have been used to improve training in many areas of classification, including digit classification~\citep{springenberg2015unsupervised, sricharan2017semi, salimans2016improved}, object classification~\citep{springenberg2015unsupervised, sricharan2017semi, salimans2016improved}, facial attribute identification~\citep{sricharan2017semi}, and image segmentation (per pixel object classification)~\citep{souly2017semi}.

\subsection{Alternative semi-supervised regression GAN methods}
\label{sec:Alternative semi-supervised regression GAN methods}
For regression, \citet{rezagholiradeh2018reg} provides two semi-supervised GAN approaches. They have applied their methods to the driving application, which we compare to in \cref{sec:Results}.

\begin{figure}
    \centering
    \includegraphics[width=0.65\columnwidth]{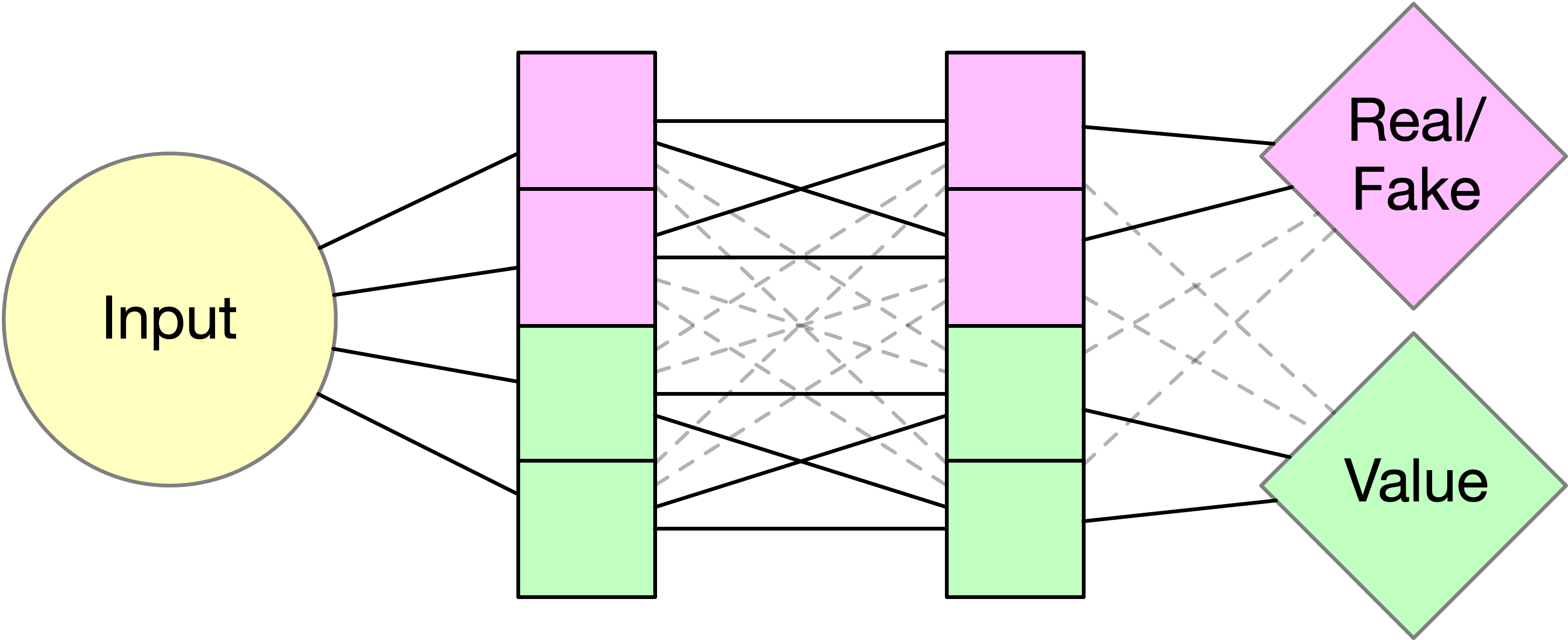}
    \caption{A DG-GAN network splitting the network into solving the two objectives independently, rather than using a shared representation. The dashed lines represent connections which exist but have very low weights. The degree of this division of learning can vary.}
    \label{fig:DGGAN network}
\end{figure}

First, they present a dual goal GAN (DG-GAN) approach, which they refer to as Reg-GAN Architecture 1. A DG-GAN outputs two labels: a regression value prediction and a fake/real classification prediction. The idea is that the network must learn both how to distinguish between real and fake examples, and how to predict the correct value for a labeled example. However, this approach does not enforce that these two predictions be related. Part of the network may learn the task of identifying real/fake images, while another portion of the network learns the task of predicting regression values. A representation of this split learning can be seen in \cref{fig:DGGAN network}. If the objective of distinguishing being real and fake examples is weighted strongly enough, the network may devote larger portions of the network to the real/fake classification task, thereby reducing its effectiveness in the regression prediction. We show in our experiments that our proposed method outperforms the DG-GAN, both in our own implementation and in that of \citet{rezagholiradeh2018reg}.

They also present a second which method, Reg-GAN Architecture 2, which only outputs the single driving angle regression value, and then attempts to label this value as fake or real depending on if the value lies within the range of real values from the dataset. This method has two significant limitations. 1) If the full range of the unlabeled dataset is unknown, a correct angle prediction will be incorrectly labeled as fake data. \citet{rezagholiradeh2018reg} assumes the range of the unlabeled data is known. 2) A bias is introduced, as values near the boundary between fake and real are preferred. This is because a generator which can exactly duplicate unlabeled data will force the discriminator to pick a value on the boundary between fake and real as the best possible answer. Finally, a discrete classification method was presented with each class being the central value of the class interval.

\subsection{Regression in Conditional GANs}
Another distinct category of related work is that of regression in conditional GANs. Conditional GANs are a type of GAN designed to produce realistic examples which have specific desired properties in the example. \citet{bazrafkan2018versatile} provides an approach to generate images with specific characteristics in a conditional GAN. In particular, they use a regressor in parallel with the discriminator network to provide more variation in the generated examples.

These works are attempting to produce realistic looking generated examples. The produce is not a predictive network for real examples. In contrast, our approach is designed to improve the predictive capabilities of the discriminator on real examples. Notably, \emph{we do not expect our generator to produce realistic looking examples}. On the contrary, we expect the examples generated will not look realistic. As noted by \citet{salimans2016improved}, the use of feature matching (which is also used in our work) improves discriminator predictive accuracy while reducing the realism of the generated examples. We expect our feature contrasting approach will further erode the realism. Furthermore, works such as \citet{dai2017good} show how a generator which produces examples that are too realistic may be less advantageous for improving a discriminator's predictive abilities.
\section{Theory and Design}
\label{sec:Theory and Design}

\subsection{SR-GAN Formulation Using Feature Contrasting}
\label{sec:SR-GAN Formulation Using Feature Contrasting}

\begin{figure}
    \centering
    \includegraphics[width=0.65\columnwidth]{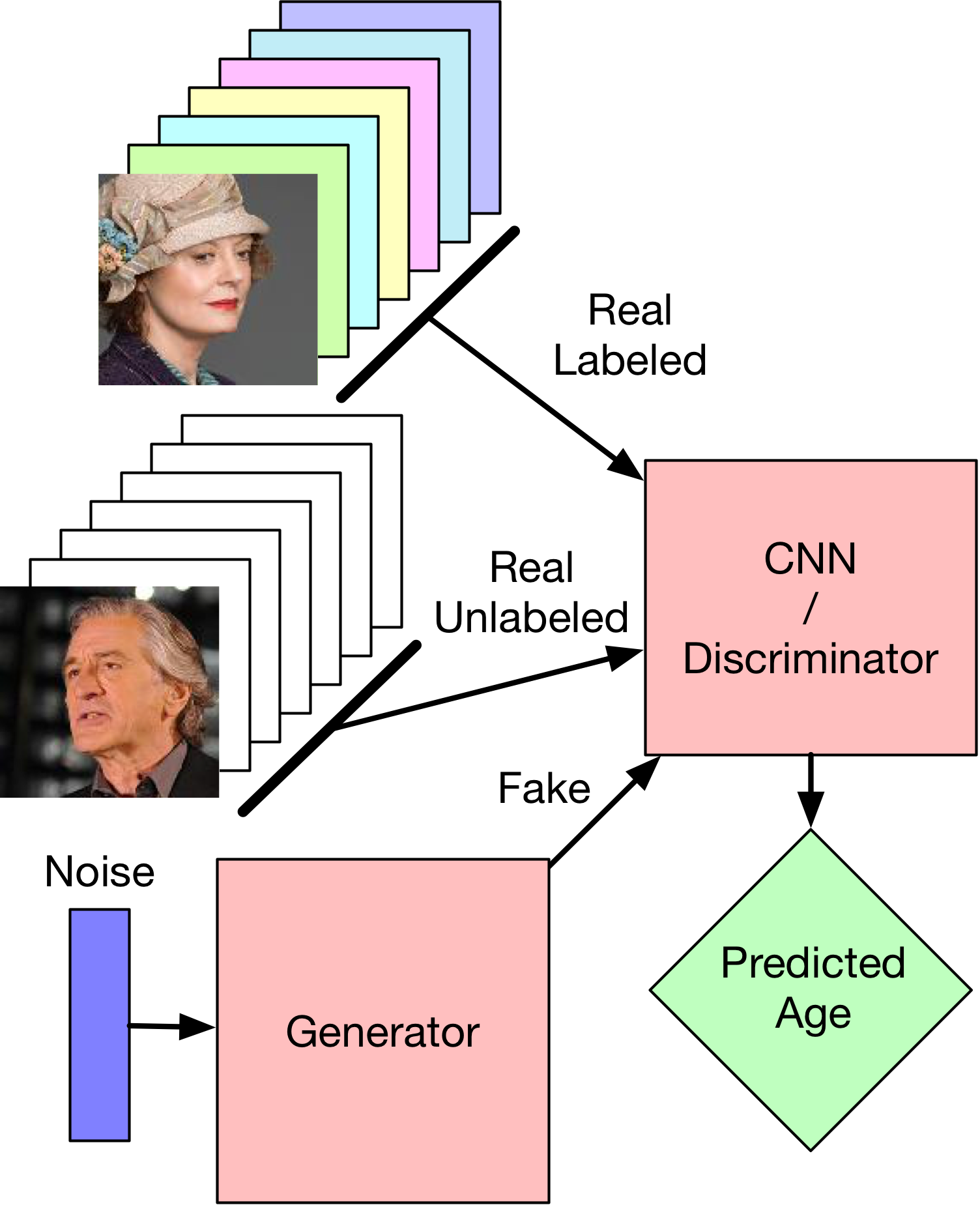}
    \caption{The structure of an SR-GAN. Its structure is similar to the semi-supervised GAN, with the major differences being in the objective functions and the output being a regression value. In this network, the discriminator distinguishes between fake and real images through feature statistics. No explicit real or fake label is assigned.}
    \label{fig:SR-GAN Diagram}
\end{figure}

The semi-supervised regression GAN (SR-GAN) approaches regression estimation by comparing the types of available data (labeled, unlabeled, and fake) as probability distributions rather than individual examples. In this method, the discriminator does not attempt to predict a label for the unlabeled data or fake data. Instead, the statistics of the features within the network for each type of data is compared. Here is the key idea: We have the discriminator seek to make the unlabeled examples have a similar feature distribution as the labeled examples. The discriminator also works to have fake examples have a feature distribution as divergent from the labeled examples distribution as possible. This forces the discriminator to see both the labeled and unlabeled examples as coming from the same distribution, and fake data as coming from a different distribution. The generator, on the other hand, will be trained to produce examples which match the unlabeled example distribution, and because of this, the generator and discriminator have opposing goals. How a label is assigned to an example drawn from that distribution is still decided by based on the labeled examples (as it is in ordinary DNN/CNN training), but the fact that the unlabeled examples must lie in the true example distribution forces the discriminator to more closely conform to the true underlying data generating distribution. The SR-GAN structure can be seen in \cref{fig:SR-GAN Diagram} with age estimation as an example. For the case of training the discriminator to have similar feature statistics for both real labeled and real unlabeled data, this approach is related to the feature matching proposed by \citet{salimans2016improved}, except that this is applied for entirely different purposes than it was in their work. In the case of training the discriminator with real data and fake data, we propose a novel approach, \emph{feature contrasting}, which is antithetical to feature matching. In this case, the discriminator attempts to make the features of the real and fake data as dissimilar as possible, while the generator is attempting to make these features as similar as possible.

\begin{figure}
    \centering
    \input{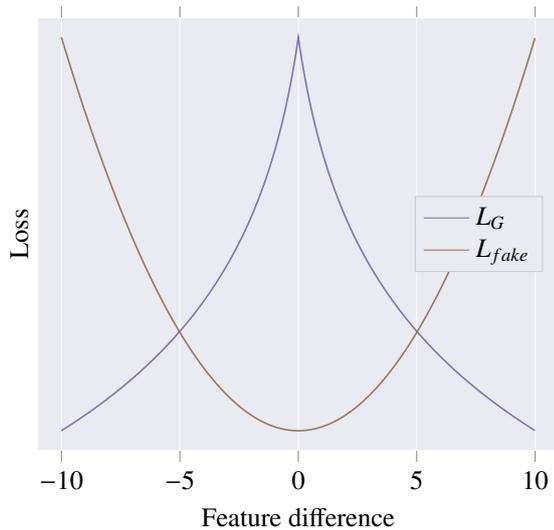}
    \caption{A comparison of the losses used for feature matching and feature contrasting, used in $L_G$ and $L_{unlabeled}$ respectively. The losses have been normalized for comparison. Shown in the change in loss due to a single feature (due to the norms used in the functions, multiple features changing together have a slightly different impact). Of particular note, a decreased loss for one necessarily results in an increased loss for the other.}
    \label{fig:SR-GAN losses}
\end{figure}

Specifically, the loss functions as defined for classification (\cref{eq:classification_discriminator_loss,eq:classification_discriminator_supervised_loss,eq:classification_discriminator_unsupervised_loss}) in the case of regression will become the following. First, we separate the loss of the discriminator into several terms for clarity. This is given by
\begin{equation}
\begin{split}
L_{D} &= L_{supervised} + L_{unsupervised} \\
&= L_{labeled} + L_{unlabeled} + L_{fake}
\end{split}\text{.}
\label{eq:discriminator_discriminator_loss}
\end{equation}
What we refer to as the "labeled loss", is given by
\begin{equation}
L_{labeled} =
\mathbb{E}_{\bm{x},y \sim p_{data}(\bm{x},y)}
[(D(\bm{x}) - y)^2]
\text{.}
\label{eq:regression_discriminator_labeled_loss}
\end{equation}
This loss is similar to an ordinary fully supervised loss (for regression training). Next, the "unlabeled loss" causes the discriminator to attempt to make the feature statistics of the real labeled data and the real unlabeled data be as similar as possible. This unlabeled loss is given by
\begin{equation}
L_{unlabeled} = \norm{
    \mathbb{E}_{\bm{x} \sim p_{labeled}} f(\bm{x}) -
    \mathbb{E}_{\bm{x} \sim p_{unlabeled}} f(\bm{x})
}^2_2
\text{.}
\label{eq:regression_discriminator_unlabeled_loss}
\end{equation}
In contrast, the "fake loss" causes to the discriminator to attempt to make the feature statistics of the real data as dissimilar to the fake data as possible. This feature contrasting is accomplished with the loss function given by
\begin{equation}
L_{fake} = - \norm{ \log \left(
    \abs*{ \mathbb{E}_{\bm{x} \sim p_{fake}} f(\bm{x}) -
    \mathbb{E}_{\bm{x} \sim p_{unlabeled}} f(\bm{x}) } + 1 \right) }_1
\text{.}
\label{eq:regression_discriminator_fake_loss}
\end{equation}
Finally, the generator attempts to make the feature statistics of the real data match those of the fake data. This goal is accomplished by the generator loss given by
\begin{equation}
L_{G} = \norm{
    \mathbb{E}_{\bm{x} \sim p_{fake}} f(\bm{x}) -
    \mathbb{E}_{\bm{x} \sim p_{unlabeled}} f(\bm{x})
}^2_2
\text{.}
\label{eq:regression_generator_loss}
\end{equation}
Here, $L_{unlabeled}$ and $L_G$ are identical except in which types of data are being compared. Additionally, the feature contrasting in \cref{eq:regression_discriminator_fake_loss} is in direct opposition to the feature matching in \cref{eq:regression_generator_loss}. Notably, there is no possibility for the generator and discriminator to both benefit by a change in these features; A decreased loss for one necessarily results an increased loss for the other. A comparison of a change in the loss from a single feature can be seen in \cref{fig:SR-GAN losses}.
We briefly explore some additional loss function options in \cref{sec:SR-GAN Formulation Using Feature Contrasting}.

We note that we choose a different norm function for \cref{eq:regression_discriminator_fake_loss} compared to \cref{eq:regression_discriminator_labeled_loss,eq:regression_generator_loss}.
The L2 norm in \cref{eq:regression_discriminator_labeled_loss,eq:regression_generator_loss} causes \emph{any} non-matching feature to be the most heavily punished, resulting in a network which tries to make all features similar. Conversely, an L1 norm is used for feature contrasting. This is because an L2 norm would result in a discriminator which focuses on the already most dissimilar feature while allowing all other features to become similar. The L1 norm puts an equal benefit on contrasting all features. To emphasize this, the L2 norm for $L_{fake}$ results in problematic backpropagation, as zero distance feature differences should result in the largest gradients, but are instead multiplied by zero.

To summarize, the SR-GAN uses feature matching for the discriminator loss functions where in previous methods a separate "fake" class is defined. Specifically this can be seen in the change from the unsupervised loss in \cref{eq:classification_discriminator_unsupervised_loss} (which uses a "fake" class in the discriminator) to \cref{eq:regression_discriminator_unlabeled_loss,eq:regression_discriminator_fake_loss} (which uses feature layer statistics). This accomplishes two goals: 
\begin{enumerate}
    \item  Regression problems have no classes and the previous methods require a "fake" class definition, and the SR-GAN approach allows regression problems to be approached.
     \item The feature matching does not introduce any bias in the discriminator label prediction, as the final label output is not used in the unsupervised loss. 
\end{enumerate}    
 Additionally, the SR-GAN approach requires no prior information about the data and requires no manual definition of goals beyond the original loss function for labeled examples.

\subsection{Gradient penalty}
\label{sec:Gradient penalty}
Of the challenges preventing the use of an SR-GAN, the greatest is likely the difficulty of designing an objective which reliably and consistently converges. GANs can easily fail to converge under various circumstances~\citet{barnett2018convergence}. To solve these general GAN instability issues, we use the gradient penalty approach proposed by \citet{arjovsky2017wasserstein} and \citet{gulrajani2017improved}.

The gradient penalty as defined by \citet{gulrajani2017improved} is not applicable to our situation, because their gradient penalty is based on the final output of the discriminator. As the final output of the discriminator is not used in producing the gradient to the generator, we use a modified form of the gradient penalty. This gradient penalty term is added to the rest of the loss function resulting in
\begin{equation}
\begin{split}
L &= L_{labeled} + L_{unlabeled} + L_{fake} \\
&+ \lambda \, \mathbb{E}_{\bm{x} \sim p_{interpolate}}\left[ \max \left( \left( \norm{\nabla_{\hat{\bm{x}}}(f(\bm{x}))}_2^2 - 1 \right), 0 \right) \right]\text{.}
\end{split}
\label{eq:gradient penalty}
\end{equation}

where $p_{interpolate}$ examples are generated by $\alpha p_{unlabeled} + (1-\alpha)p_{fake}$ for $\alpha \sim \mathcal{U}$. The last term basically provides a restriction on how quickly the discriminator can change relative to the generator's output. Our version of the gradient penalty term is modified in multiple ways from the original. First, as noted above, the final discriminator output cannot be used, nor should it, as the discriminator's interpretation of the generated data only matter in regard to the feature vector, $f(\bm{x})$. Second, the gradient penalty is normally applied to a term similar to the $L_{fake}$ term using the interpolated values. However, our $L_{fake}$ is based on the average of a batch of fake examples whose difference is then taken from a batch of real examples. As both the $L_{fake}$ term and interpolates are calculated based on the real data, the resulting gradient penalty is negligible. Instead, we apply the gradient penalty directly to the mean feature vector of a batch of interpolated examples and do not apply the feature distance loss function compared to the mean real feature vector. As this penalizes the gradient even for mean feature vectors far from the mean real feature vector, it may slow training. However, near the real feature vector, it approximates the original gradient penalty formulation and works well in practice. Lastly, we use the one-sided version of the gradient penalty described by \citet{gulrajani2017improved}. As mentioned in their work, the one-sided penalty more closely matches the desired discriminator training properties, and we found this approach to produce higher accuracies than the two-sided penalty.

\section{Experiments and Results}
\label{sec:Results}

To demonstrate the capabilities of the semi-supervised regression GANs, we use four experimental setups, each of which consists of several individual trials and demonstrations.


The first experimental setup will be of a synthesized dataset problem. This will allow us to demonstrate the details of the theoretical issues behind a semi-supervised regression GAN in a well controlled and understood environment. These include: what is the right objective which reliably and consistently converges in training, and how little data is needed to achieve different levels of prediction accuracy.  We will use a dataset of polynomials with sampled points on the polynomial, whereas the goal of the network is to predict coefficients of the polynomial given the sampled points. Using this simplistic problem, we can show how the semi-supervised regression GAN works in details, what variations can influence its capabilities, and what its limitations are. Most importantly, this allows us to have complete control and understanding of the underlying data generating distribution. This is impossible in any real-world application, as the underlying data generating distribution there is the real world itself.


The downside to the synthetic dataset is that because we have complete control over the data generating distribution, we can define the data such that our SR-GAN does arbitrarily well compared with a normal DNN. As such, the remaining experimental setups are real-world applications. The applications of age estimation, driving steering angle prediction, and crowd counting have been chosen for this purpose. The real world case provides an area we can show direct improvements in compared to a non-adversarial CNN.

\subsection{Coefficient Estimation}

The first experimental setup consists of a simple, well-controlled mathematical model, whose problem can be easily solved with simple neural networks when given enough examples. The example chosen is a polynomial coefficient estimation problem. This problem allows for an environment in which many properties of the semi-supervised regression GAN can be shown and their limits tested. In particular, the simple environment allows us to not only demonstrate the properties of the semi-supervised regression GAN but also give a clear theoretic understanding of why the network exhibits these behaviors. Five important aspects will be discussed: 1) the dataset; 2) the experiment setup; 3) estimation with minimal data; 4) loss function analysis; and 5) choices of gradient penalty.

\subsubsection{Polynomial Coefficient Estimation Dataset}
\label{sec:mathematical_model}

For the data of the mathematical model to appropriately represent the characteristics of a real aggression application, we seek to create a data generating model that exhibits the following properties.

\begin{enumerate}
    \item Able to produce any desired number of examples. \label{item:unlimited_examples} 
    \item The distribution of the underlying data properties is selectable. \label{item:underlying_distribution}
    \item The relation between the raw data and the label is abstract, where the label is a regression value instead of one of a finite number of classes. \label{item:abstract_relationship}
    \item Able to contain latent properties that effect the relation between the data and the labels. \label{item:latent_properties}
    \item Most of the data can be made to be irrelevant to the label. \label{item:irrelevant_information}
\end{enumerate}

Property 1 allows us to run any number of trials on new data, and run trials where data is unlimited. Property 2 reveals the inner workings of the data distribution. This is important, as we can monitor how closely the generator's examples match the true distribution and examine what kinds of distributions lead to limitations or advantages of the GAN model. Property 3 ensures the findings on the toy model is relevant real deep learning applications for regression. That is, deep learning is typically used in cases where input data is complex, and an abstract, high-level meaning of that data is desired. When the relationship between the data and the label (the regression value) is too simple, more traditional prediction methods tend to be used. Property 4 is also important because of its relationship to real applications. Most applications involve cases where a property which is not the value to be predicted directly effects the data related to value to be predicted. For example, in the case of age estimation, whether the image of the face is lit from the front or lit from the side drastically changes the data and what the CNN should be searching for. Finally, Property 5 requires that our model is able to filter which pieces of information are important and which are not. Again, in the case of age estimation, whether the background behind the person is outdoors or indoors should have little or no impact on the prediction of their age. In many, if not most, cases of deep learning applications the majority of the input data has little to no relevance for the task at hand. The network must learn which information should be relied on and which data should be ignored.

An option of a simplistic mathematical model for this purpose would be a data generating distribution which is defined as follows. First, we define a polynomial,

\begin{equation}
y = a_4 x^{4} + a_3 x^{3} + a_2 x^{2} + a_1 x
\text{.}
\label{eq:polynomial model}
\end{equation}

\begin{figure}
    \centering
    \input{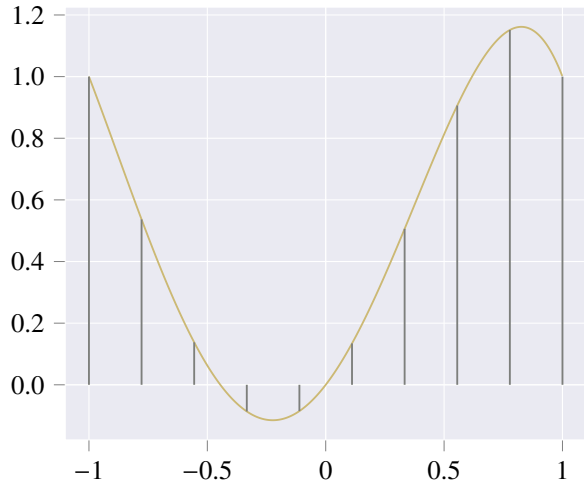}
    \caption{An example of a polynomial as described in \cref{eq:polynomial model} with 10 points sampled. In this case, $a_2 = 2$, $a_3 = -1$, and $a_4 = -1$, but only $a_3$ is the coefficient to be estimated.}
    \label{fig:example polynomial}
\end{figure}

We set $a_1 = 1$. With $\mathcal{U}(r_0, r_1)$ representing a uniform distribution over the range from $r_0$ to $r_1$, $a_3$ is randomly chosen from $\mathcal{U}(-1, 1)$. $a_2$ and $a_4$ are randomly chosen from $b\cdot\mathcal{U}(-2, -1) + (1-b)\cdot\mathcal{U}(1, 2)$ with $b$ being randomly chosen from a standard binomial distribution. Then we sample $y$ for 10 $x$s from linear space from $-1$ to $1$. An example of such a polynomial and the observed points are shown in \cref{fig:example polynomial}. This one polynomial and the observed points constitutes a single example in our dataset. The label of this example we choose as $a_3$. That is, our network, when given the 10 observations, should be able to predict $a_3$.

We can compare the pieces of this data generating distribution to the standard image regression problem (think of age estimation from images) to better understand what parts of the toy model represent which parts in a real application model. The 10 observed values from the toy model are analogous to the pixel values in image regression. $a_3$ is equivalent to the object label (e.g. age value). Finally, the set of all polynomials obtainable from \cref{eq:polynomial model}, given the restrictions on how the coefficients are chosen, is the underlying data generating distribution in the toy case, where this role is played by views of the real world projected to an image plane in the regression case (such as age estimation).

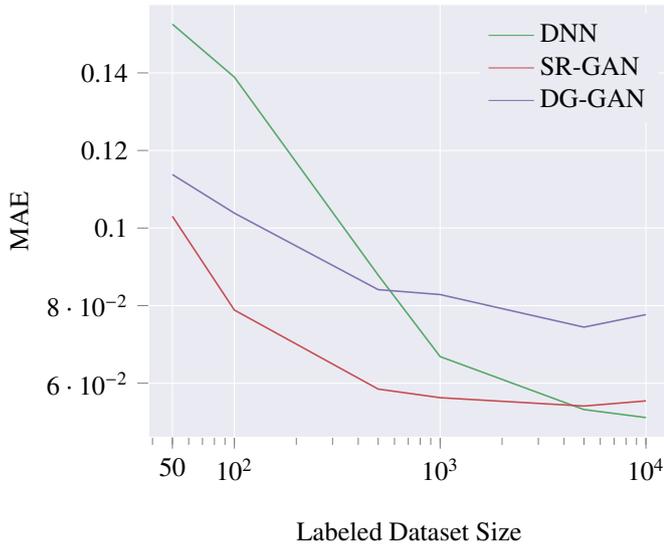
\begin{figure}
    \centering
\begin{tikzpicture}

\definecolor{color0}{rgb}{0.917647058823529,0.917647058823529,0.949019607843137}
\definecolor{color2}{rgb}{0.768627450980392,0.305882352941176,0.32156862745098}
\definecolor{color1}{rgb}{0.333333333333333,0.658823529411765,0.407843137254902}
\definecolor{color3}{rgb}{0.505882352941176,0.447058823529412,0.701960784313725}

\begin{axis}[
xlabel={Labeled Dataset Size},
ylabel={MAE},
xmin=38.3635249505462, xmax=13033.2132056306,
ymin=0.0460618643388152, ymax=0.157625743158162,
xmode=log,
tick align=outside,
tick pos=left,
xmajorgrids,
x grid style={white},
ymajorgrids,
y grid style={white},
axis line style={white},
axis background/.style={fill=color0},
legend entries={{DNN},{SR-GAN},{DG-GAN}},
legend cell align={left},
legend style={draw=none, fill=color0},
extra x ticks={50},
extra x tick style = {
    log identify minor tick positions=false,
    log ticks with fixed point,
}
]
\addlegendimage{no markers, color1}
\addlegendimage{no markers, color2}
\addlegendimage{no markers, color3}
\addplot [semithick, color1]
table [row sep=\\]{%
50	0.152554657757282 \\
100	0.138916473339001 \\
500	0.0878596296906471 \\
1000	0.066851079761982 \\
5000	0.0531993109732866 \\
10000	0.0511329497396946 \\
};
\addplot [semithick, color2]
table [row sep=\\]{%
50	0.103021762222052 \\
100	0.0788566852609316 \\
500	0.058481405004859 \\
1000	0.056259433105588 \\
5000	0.0541069766134024 \\
10000	0.0554335790872574 \\
};
\addplot [semithick, color3]
table [row sep=\\]{%
50	0.113793955378899 \\
100	0.103840945783853 \\
500	0.0841209090839786 \\
1000	0.0828509258803522 \\
5000	0.0744501822230252 \\
10000	0.0776832339960312 \\
};
\end{axis}

\end{tikzpicture}
    \caption{The resultant inference accuracy of the coefficient estimation network trained with and without the SR-GAN for various quantities of labeled data.}
    \label{fig:coef-gan-vs-dnn}
\end{figure}
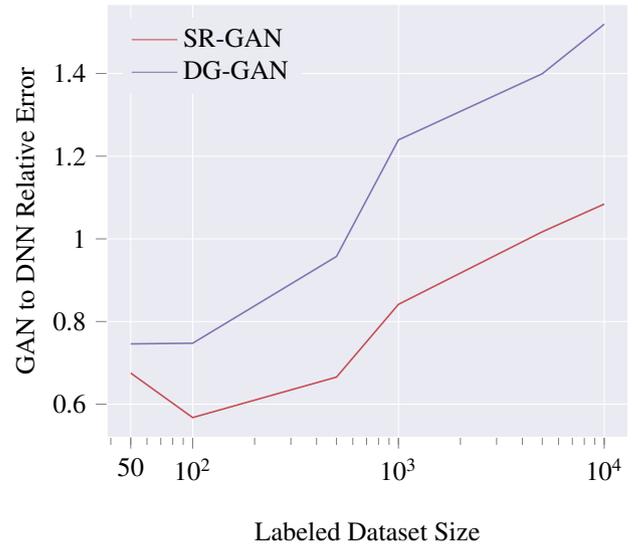
\begin{figure}
    \centering
\begin{tikzpicture}

\definecolor{color0}{rgb}{0.917647058823529,0.917647058823529,0.949019607843137}
\definecolor{color1}{rgb}{0.768627450980392,0.305882352941176,0.32156862745098}
\definecolor{color2}{rgb}{0.505882352941176,0.447058823529412,0.701960784313725}

\begin{axis}[
xlabel={Labeled Dataset Size},
ylabel={GAN to DNN Relative Error},
xmin=38.3635249505462, xmax=13033.2132056306,
ymin=0.520076151420755, ymax=1.56681946513583,
xmode=log,
tick align=outside,
tick pos=left,
xmajorgrids,
x grid style={white},
ymajorgrids,
y grid style={white},
axis line style={white},
axis background/.style={fill=color0},
legend entries={{SR-GAN},{DG-GAN}},
legend cell align={left},
legend style={draw=none, fill=color0, at={(0.03,0.97)}, anchor=north west},
extra x ticks={50},
extra x tick style = {
    log identify minor tick positions=false,
    log ticks with fixed point,
}
]
\addlegendimage{no markers, color1}
\addlegendimage{no markers, color2}
\addplot [semithick, color1, forget plot]
table [row sep=\\]{%
50	0.675310500096047 \\
100	0.567655392953258 \\
500	0.665623167440739 \\
1000	0.841563566450912 \\
5000	1.0170616051883 \\
10000	1.08410681115516 \\
};
\addplot [semithick, color2, forget plot]
table [row sep=\\]{%
50	0.745922524109018 \\
100	0.747506348872298 \\
500	0.957446660999682 \\
1000	1.23933564237611 \\
5000	1.39945764072789 \\
10000	1.51924022360333 \\
};
\end{axis}

\end{tikzpicture}
    \caption{The relative error of the GAN model over CNN model for various quantities of labeled data for the coefficient model.}
    \label{fig:coef-gan-to-dnn-relative-error}
\end{figure}

This model fulfills all but the last property defined above. To satisfy Property 5, we simply make every example in the dataset consist of 5 different polynomials each chosen and observed as previously explained. However, for this single example (consisting of 5 polynomials) on the $a_3$ coefficient of the first example is the label. Thus, each example consists of 50 observations, only 10 of which are related to the label. Lastly, we apply noise to every observation.
\subsubsection{Coefficient Estimation Experimental Setup}
In the coefficient estimation experiments, both the discriminator and generator each consisted of a 4 layer fully connected neural network. Each layer contained 10 hidden units. All code and hyperparameters can be found at \url{https://github.com/golmschenk/srgan}. The training dataset for each experiment was randomly chosen. The seed for the random number generator is set to 0 for the first experiment, 1 for the second, and so on. The same seeds are used for each set of experiments. That is, the SR-GAN compared with the DNN use the same training data for each individual trial. Additionally, for experiments over a changing hyperparameter, the same seeds are used for each hyperparameter value.

In these experiments, we demonstrate the value of the SR-GAN on polynomial coefficient estimation. Using a simple fully connected neural network architecture, we have tested the DG-GAN and SR-GAN methods compared to a plain DNN on various quantities of data from the generation process described above. The results of these experiments can be seen in \cref{fig:coef-gan-vs-dnn}. In each of these experiments an unlabeled dataset of 50,000 examples was used, when various quantities (from 50 to 10,000) of labeled data were used. Each data point on the plots is the average of three training runs randomly seeded to contain different training and test sets on each experiment. The relative error between the DNN and the GAN methods can be seen in \cref{fig:coef-gan-to-dnn-relative-error}. We see a significant accuracy improvement in lower labeled data cases for the GAN methods. The SR-GAN error is 68\% of what the DNN error is at with 50 labeled examples. With 50 examples, the DG-GAN also has a significant advantage with 75\% the error the DNN has. However, the DG-GAN quickly loses its advantage over the DNN as the data size increases.  As the amount of labeled data becomes very large, SR-GAN does not perform better than the DNN. This diminishing return is expected, as we can consider the case of infinite labeled data, where unlabeled data could then provide no additional useful information. We note that for the simple problem of coefficient estimation, 10,000 examples is a very large dataset for training. In each of the real world applications we tested our SR-GAN method in, we did not see a detriment in using the SR-GAN with larger numbers of labeled examples.

\subsubsection{Loss Function Analysis on Coefficient Estimation}
\label{sec:Loss Functional Analysis}
As noted in \cref{sec:SR-GAN Formulation Using Feature Contrasting} we primarily experimented with the loss functions given in \cref{eq:regression_discriminator_fake_loss,eq:regression_discriminator_labeled_loss,eq:regression_generator_loss}. However, these are not the only loss functions which could be used for the feature matching and feature contrasting objectives.

We tested three sets of loss functions. We will refer to the feature distance vector as
\begin{equation}
    \bm{d}_f = \abs*{\mathbb{E}_{\bm{x} \sim p_{1}} f(\bm{x}) -
    \mathbb{E}_{\bm{x} \sim p_{2}} f(\bm{x})}
\end{equation}
where $p_1$ and $p_2$ are the appropriate labeled, unlabeled, or fake data distributions depending on if the $d_f$ is being used in the $L_{unlabeled}$, $L_{fake}$, or $L_G$
 terms. With this, we used the feature contrasting and feature matching loss functions given in \cref{tab:Loss Order Analysis Table}. The first is the set of loss functions given previously, which we have already given an explanation for. The second set keeps the same feature matching function but uses a square root as the primary component of the feature contrasting function. This provides a stronger incentive for the discriminator to push features which are already far apart, even further apart. This second approach did slightly worse than the first, likely because focusing on contrasting those features which are most similar between the fake and real examples provides a greater improvement. The third approach uses linear losses. This is similar to the linear fake/real losses used in the WGAN implementation by \citet{arjovsky2017wasserstein}. The reason for the decreased accuracy is likely the same as for the second case, where features which are already dissimilar are still given too much priority in the feature contrasting.
\begin{table}
    \centering
    \begin{tabular}{lr}
        \toprule
        Loss Functions & MAE \\
        \midrule
        $L_{fake} = - \norm{ \log \left( \bm{d}_f + 1 \right) }_1\quad L_{unlabeled}=L_{G}=\norm{\bm{d}_f}^2_2$
        & 0.0578 \\
        $L_{fake} = - \norm{ \sqrt{ \bm{d}_f + 1} }_1\quad L_{unlabeled}=L_{G}=\norm{\bm{d}_f}^2_2$
        & 0.0613 \\
        $L_{fake} = -\norm{\bm{d}_f}_1 \quad L_{unlabeled}=L_{G}=\norm{\bm{d}_f}_2$
        & 0.0672 \\
        \bottomrule
    \end{tabular}
    \caption{A comparison of the SR-GAN method using various loss functions for feature matching and feature contrasting. Each experiment was run on the coefficient application with 500 labeled examples and 50,000 unlabeled examples.}
    \label{tab:Loss Order Analysis Table}
\end{table}

\subsection{Driving Steering Angle Prediction}
This application works to predict the steering angle of a car given an image from the front of a car. Such an approach allows for basic partial self-driving/auto-pilot capabilities using a single image~\citep{pan2017virtual}. The dataset~\citep{chen2017driving} consists of 45,567 images from a dashboard-mounted camera, where for each image the current rotation angle of the steering wheel was recorded. The goal of the network is to predict this rotation angle given the front facing view image, whose primary feature is the upcoming road segment.

\citet{rezagholiradeh2018reg} provides two semi-supervised GAN approaches to train for this application which are described in \cref{sec:Alternative semi-supervised regression GAN methods}. Additionally, they also provide a baseline discrete classification method with each class being the central value of the class interval.

\begin{figure}
    \centering
    \includegraphics[width=0.5\columnwidth]{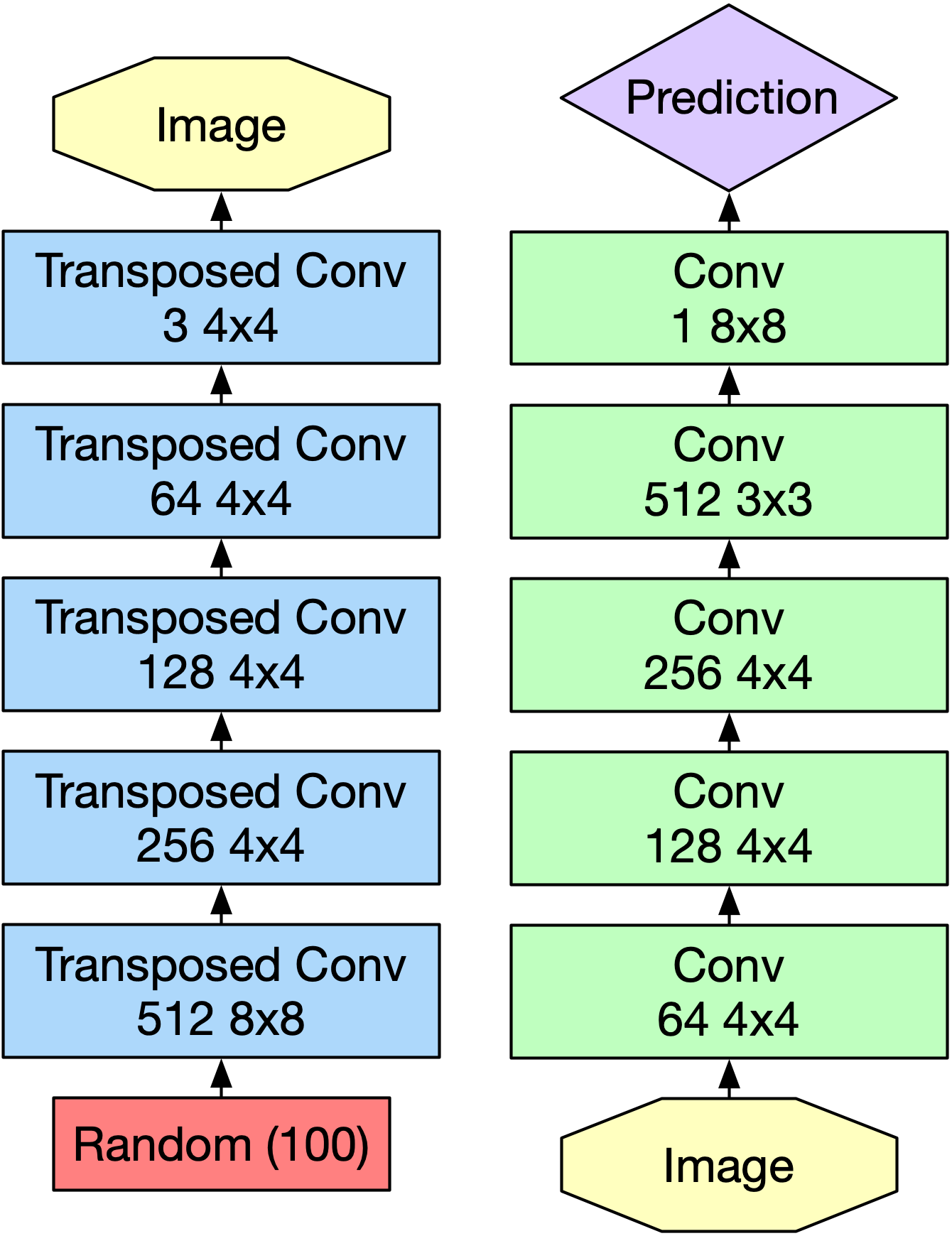}
    \caption{The DCGAN structure used for the age estimation experiments. The left network is the generator and the right is the discriminator/CNN.}
    \label{fig:Age Network}
\end{figure}

\begin{figure}
\centering
\begin{subfigure}{.5\linewidth}
  \centering
  \includegraphics[width=.95\linewidth]{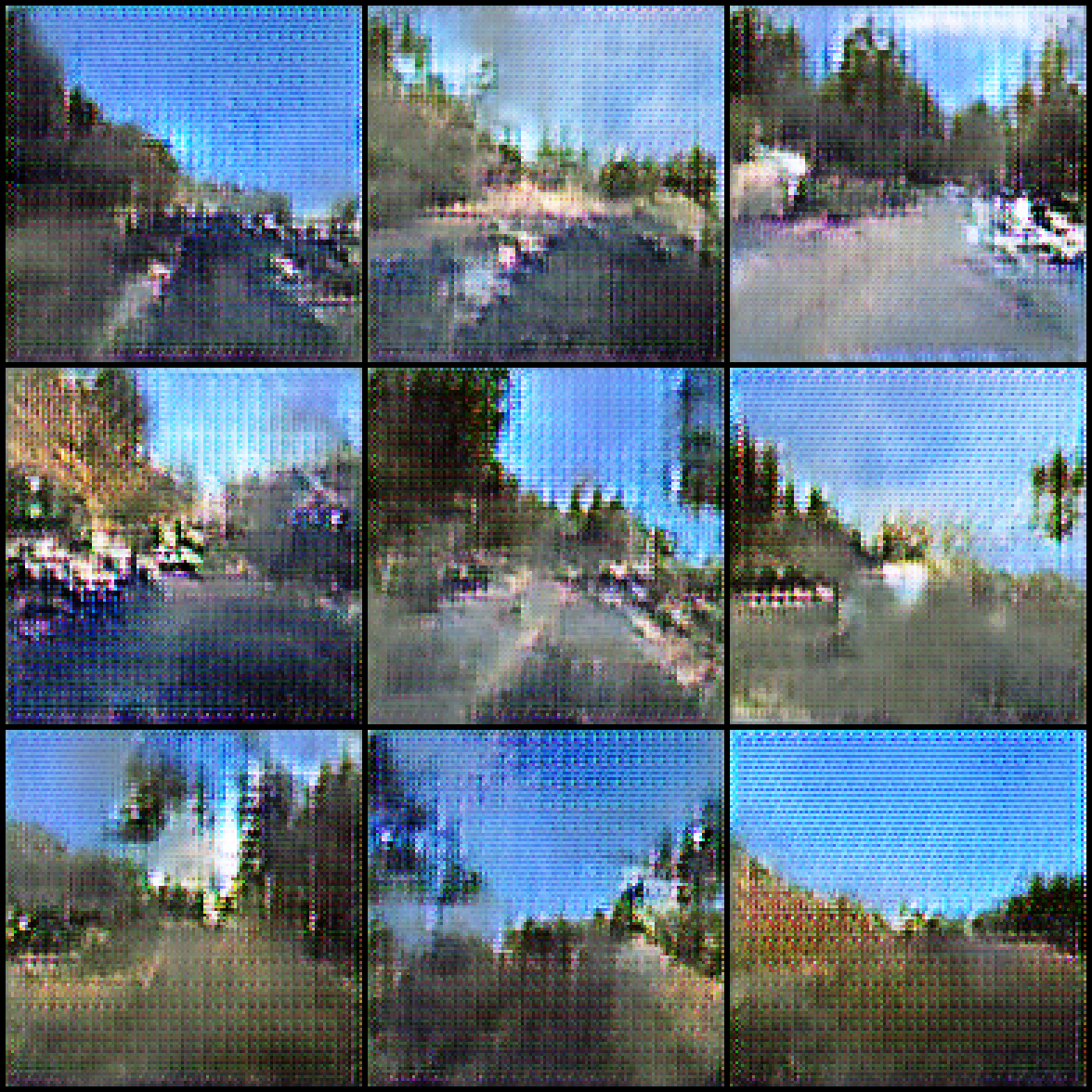}
  \caption{Fake steering angle images.}
\end{subfigure}%
\begin{subfigure}{.5\linewidth}
  \centering
  \includegraphics[width=.95\linewidth]{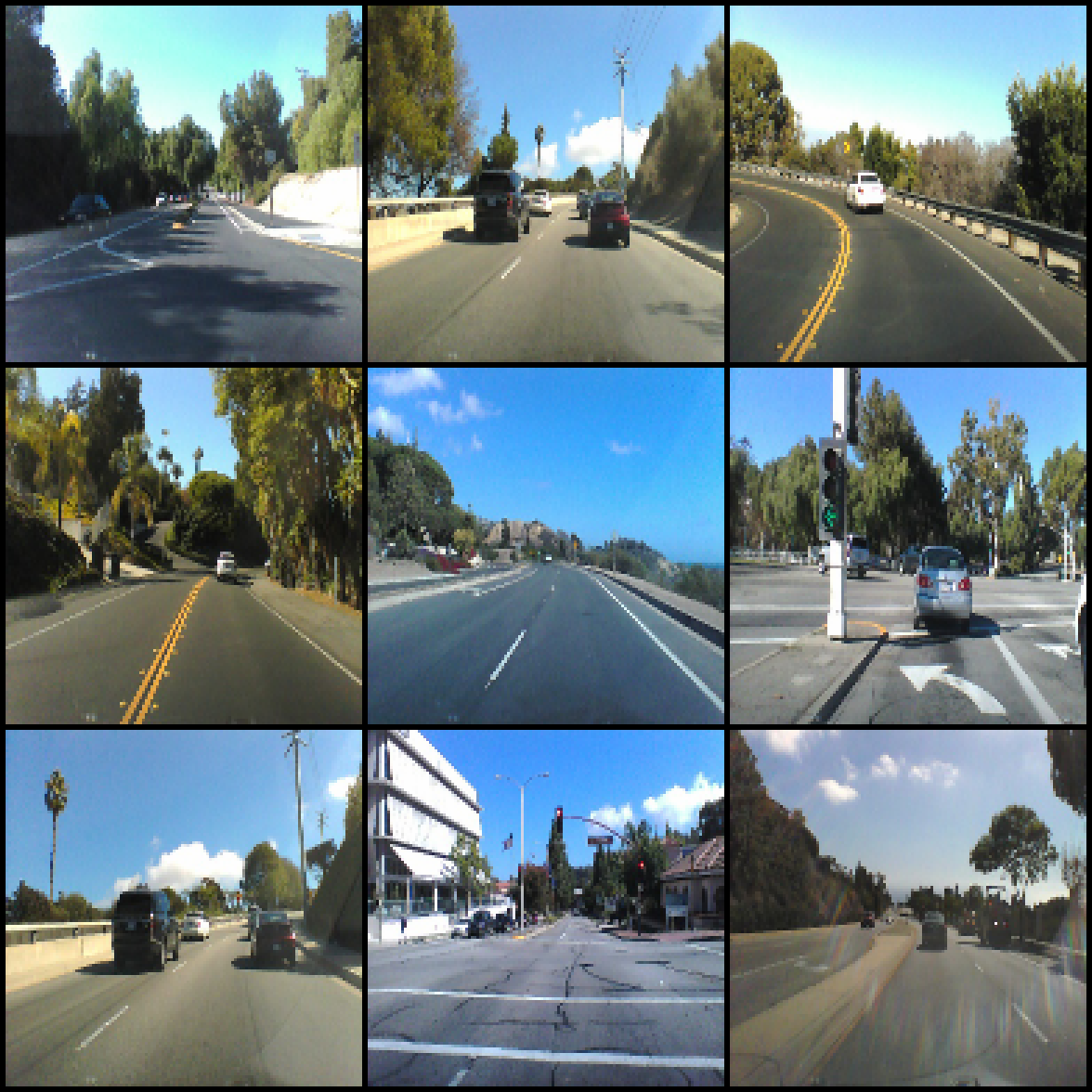}
  \caption{Real steering angle images.}
\end{subfigure}
\caption{Examples of real and fake images used/generated during training. We note that our approach is not intended to produce realistic looking images, and the fake images are only included for insight.}
\label{fig:Driving examples}
\end{figure}

Here, we perform the experiments presented by \citet{rezagholiradeh2018reg} using our SR-GAN approach. In these experiments, varying numbers of labeled images randomly selected from the entire dataset are used for training (up to 7,200 images) and testing (9,000 images). The remaining images are used as the unlabeled data. We use the DCGAN network architecture~\citep{radford2015unsupervised}, which matches the architecture presented by \citet{rezagholiradeh2018reg}. This network structure (both generator and discriminator) is shown in \cref{fig:Age Network}. All code and hyperparameters can be found at \url{https://github.com/golmschenk/srgan}. We note that we cannot precisely duplicate the experiments by \citet{rezagholiradeh2018reg}, as the images used for training and testing were randomly chosen. We similarly randomly selected our datasets. Our random selections were seeded for reproducibility, and the code at our repository can be used to retrieve the dataset selection for our experiments. Examples of the images, both real and fake, used/generated during training are shown in \cref{fig:Driving examples}.

We also note that an entirely random image selection has limited evaluation value for this dataset. The images are part of a video sequence with each image have only minor differences from the previous image. Even a small percentage of the images, when randomly chosen, will contain the primary attributes of a large portion of the dataset. However, for comparison purposes, we have followed the experimental procedure used by \citet{rezagholiradeh2018reg}. We have additionally provided results for significantly lower numbers of labeled images.

The evaluation metric used is a normalized mean absolute error (NAE) given by
\begin{equation}
    NAE = \dfrac{1}{N}\sum^N_{i=1}\dfrac{\abs{y_i - \hat{y}_i}}{y_{max} - y_{min}} \times 100\%\text{.}
\end{equation}

\begin{table*}
    \centering
    \begin{tabular}{lrrrrrr}  
        \toprule
        Method & 100 & 500 & 1000 & 2000 & 4000 & 7200 \\
        \midrule
        Improved-GAN & - & - & 4.38\% & 4.22\% & 4.07\% & 4.06\% \\
        Reg-GAN (Arch 1) & - & - & 2.43\% & 2.40\% & 2.39\% & 2.36\% \\
        Reg-GAN (Arch 2) & - & - & 3.81\% & 3.58\% & 2.23\% & 2.21\% \\
        \midrule
        \textbf{SR-GAN} & 3.12\% & 2.32\% & 2.02\% & 1.89\% & 1.37\% & 1.16\% \\
        \bottomrule
    \end{tabular}
    \caption{Steering angle prediction NAE compared to existing approaches for various amounts of labeled training examples.}
    \label{tab:Driving results}
\end{table*}

The results of our method in comparison to the methods presented by \citet{rezagholiradeh2018reg} are shown in \cref{tab:Driving results}. In these experiments, we show that our SR-GAN method significantly outperforms the Reg-GAN method for any number of labeled examples. As Architecture 2 is the more generalized approach of Reg-GAN, it provides the comparison of the most interest.

\subsection{Age Estimation}
Age estimation is a well-known regression problem in computer vision using deep learning. In particular, well-established datasets of images of individuals with corresponding ages exist and are widely used by the computer vision community. The most notable age estimation database is currently the IMDB-WIKI Face Dataset~\citep{rothe2016deep}.

For our work, having such a well-known dataset is particularly important as the deep learning community tends to focus on classification problems and not regression problems. Due to this, well-known regression datasets---ones known even outside their domain---tend to be rare. The age estimation dataset is one of these rare cases. It provides a standard which we can test our SR-GAN on which is widely tested on.

\subsubsection{Age Estimation Dataset}
The IMDB-WIKI dataset includes over 0.5 million annotated images of faces and the corresponding ages of the people thus imaged. There are 523,051 face images: from 20,284 celebrities, 460,723 face images are from IMDb and 62,328 from Wikipedia. 5\% of the celebrities have more than 100 photos, and on average each celebrity has around 23 images.

There are likely many mislabeled images included in the dataset. The image-label pairs were created by searching the top 100,000 actors on IMDb (also known as the "Internet Movie Database). The actors' IMDb profile and Wikipedia page were scraped for images. Face detection was performed on these images, and if a single face detection is found, the image is assumed to be of the correct individual. The image timestamp along with the date of birth of the actor is used to label the image with an age. The image is often a screen capture of a movie, which may have taken years to produce or the screen capture may have happened years later. Additionally, the actor may be purposely made to look a different age in the movie. Despite these many areas of mislabeling, the dataset it thought to consist of overwhelmingly correctly labeled images. To minimize the number of incorrectly labeled images the database is filtered based on several criteria. The database includes face detection scores (certainty of containing a face) and a secondary face score (containing an additional face). If the first face score was too low the image was excluded. If there was a secondary face detected it is also excluded (since these are taken from the actor's IMDb page, it is only assumed to be a picture of the actor if there is only one person in the image). Images labeled with an age below 10 or above 95 are also excluded. Primarily, the below 10 filter is important as many images included an incorrect age of only a few years old. Finally, only images of 256\texttimes256 resolution or higher are used. After this filtering, we are left with \approximately90K images. Both the labeled and unlabeled data is taken from these images (without overlap), and the labels were not used for the unlabeled data. Data was selected randomly for each trial. Though other face data could be used for the unlabeled data, for these experiments, we wished to ensure that the labeled and unlabeled data came from the same data distribution.
\begin{figure}
\centering
\begin{subfigure}{.5\linewidth}
  \centering
  \includegraphics[width=.95\linewidth]{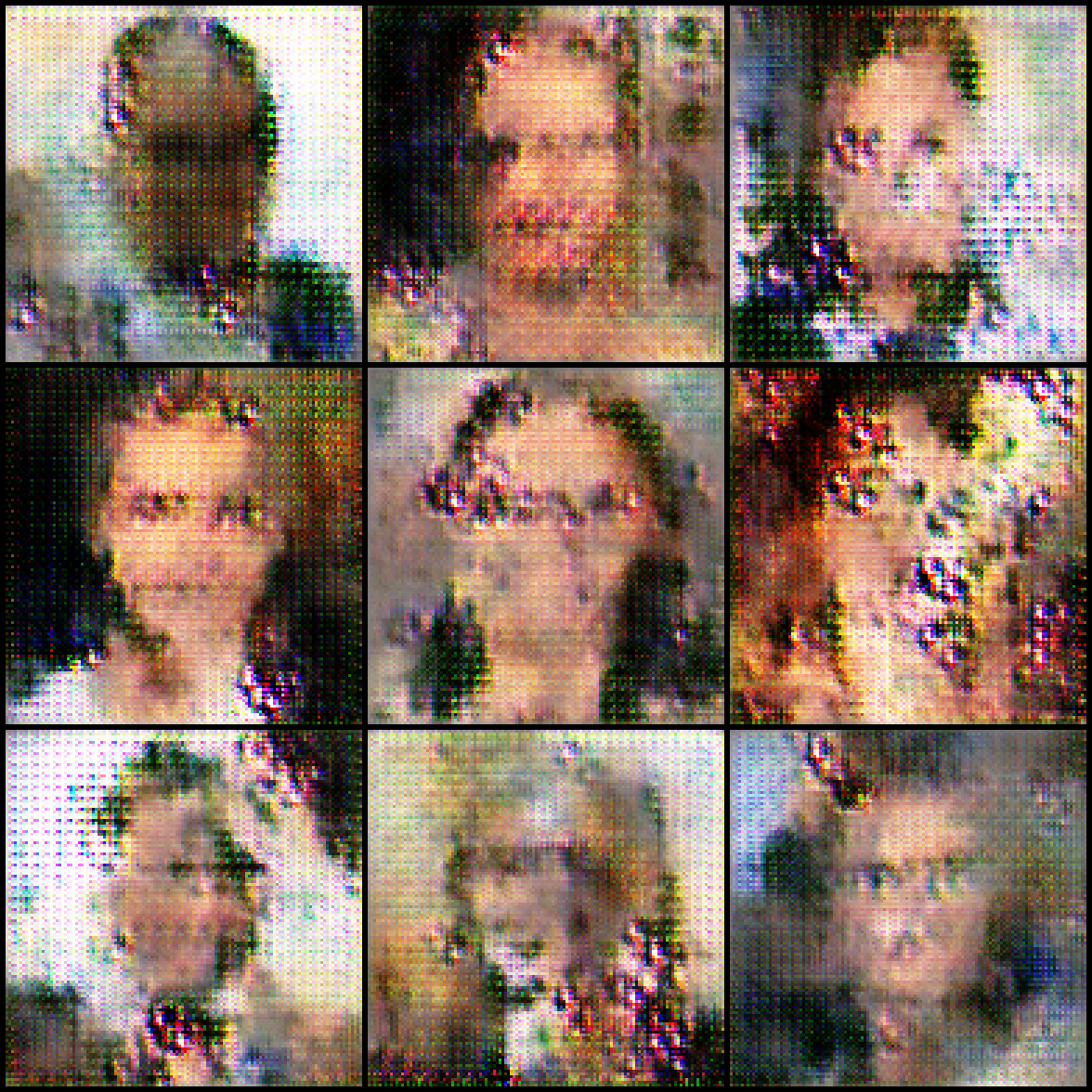}
  \caption{Fake age images.}
\end{subfigure}%
\begin{subfigure}{.5\linewidth}
  \centering
  \includegraphics[width=.95\linewidth]{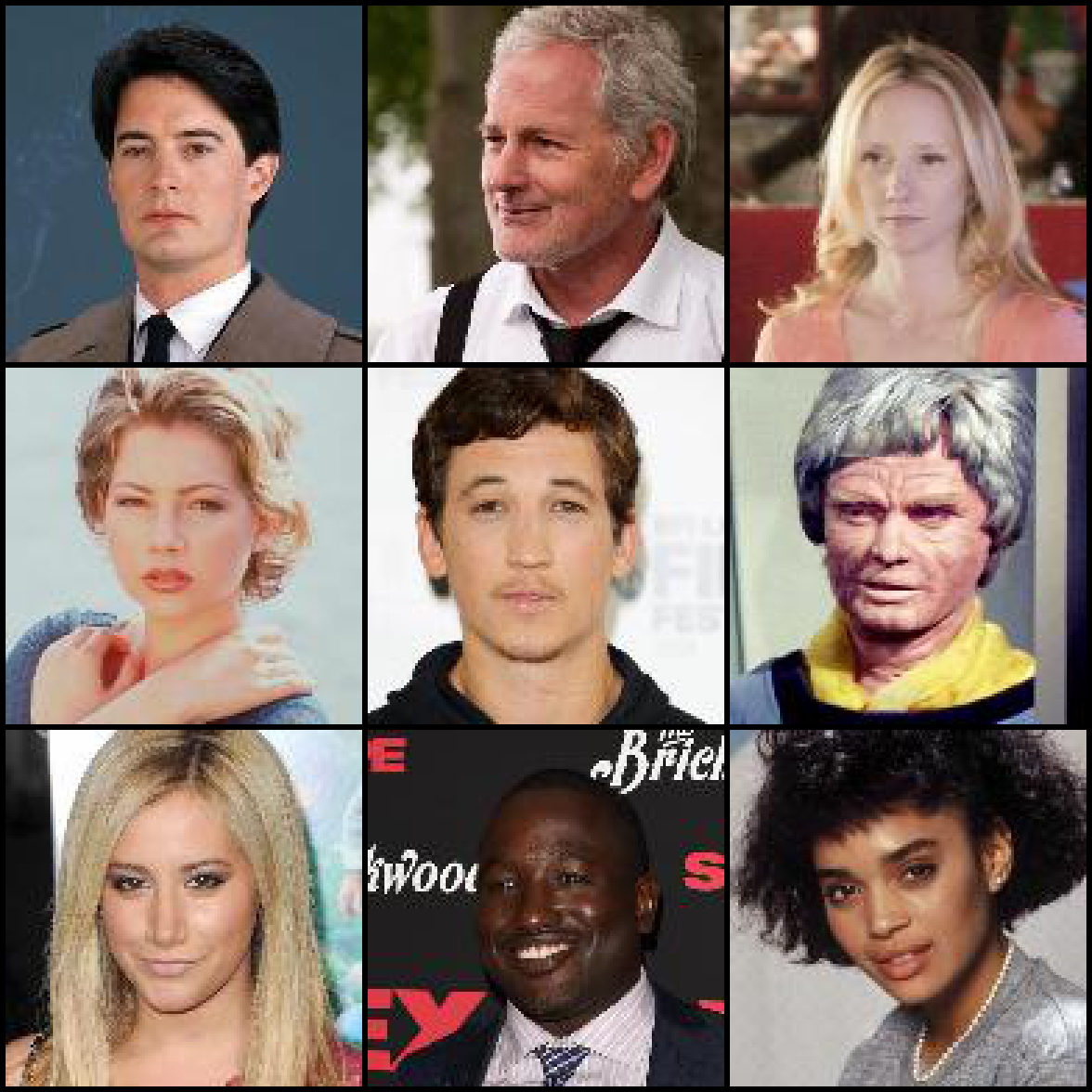}
  \caption{Real age images.}
\end{subfigure}
\caption{Examples of real and fake images used/generated during training. We note that our approach is not intended to produce realistic looking images, and the fake images are only included for insight.}
\label{fig:Age Examples}
\end{figure}
\begin{figure}
    \centering
    \input{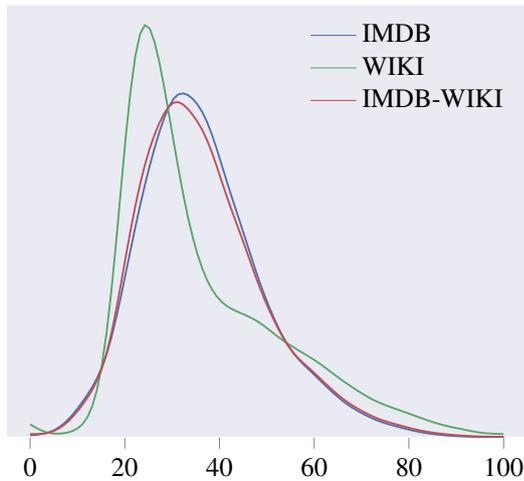}
    \caption{The distribution of ages in the IMDB-WIKI database.}
    \label{fig:age-distribution}
\end{figure}

\subsubsection{Age Estimation Experimental Setup}
In the age estimation experiments, the DCGAN network architecture~\citep{radford2015unsupervised} is used. All code and hyperparameters can be found at \url{https://github.com/golmschenk/srgan}. The discriminator of the DCGAN was used alone as the CNN baseline model. The network structure can be seen in \cref{fig:Age Network}. The training dataset for each experiment was randomly chosen. The seed is set to 0 for the first experiment, 1 for the second, and so on. The same seeds are used for each set of experiments. That is, the SR-GAN compared with the CNN use the same training data for each individual trial. This set of experiments used the second set of loss functions from \cref{sec:Loss Functional Analysis}.

\begin{figure}
    \centering
    \input{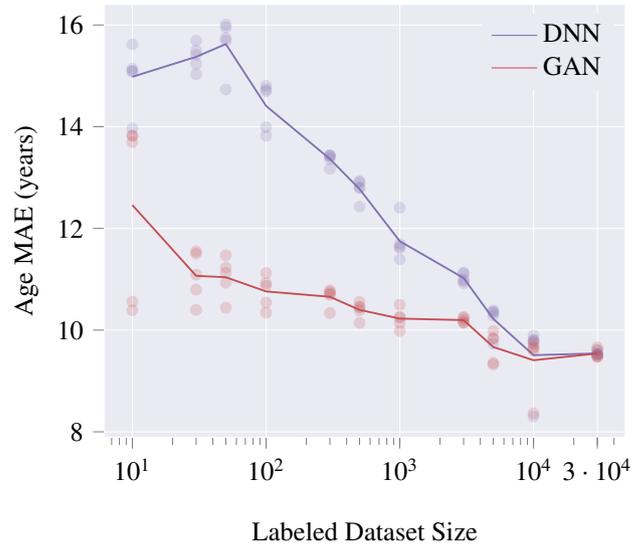}
    \caption{The resultant inference accuracy of the age estimation network trained with and without the SR-GAN for various quantities of labeled data. Each dot represents a trial with randomized training data, and the line represents the mean of the trials.}
    \label{fig:age-systematic-gan-vs-dnn}
\end{figure}

\begin{figure}
    \centering
    \input{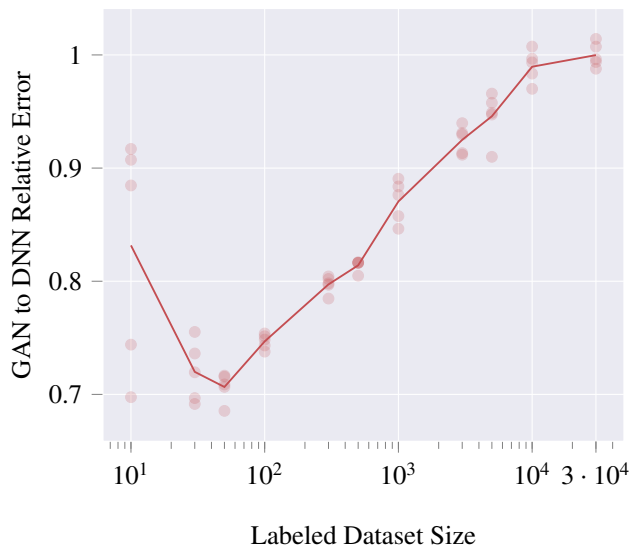}
    \caption{The relative error of the GAN model over CNN model for various quantities of labeled data for age estimation. Each dot represents a trial with randomized training data, and the line represents the mean of the trials.}
    \label{fig:age-systematic-gan-to-dnn-relative-error}
\end{figure}

The following experiments demonstrate the value of the SR-GAN on age estimation. Using a DCGAN~\citep{radford2015unsupervised} network architecture, we have tested the SR-GAN method on various quantities of data from the IMDB-WIKI database. The results of these experiments can be seen in \cref{fig:age-systematic-gan-vs-dnn}. In each of these experiments, an unlabeled dataset of 50,000 images was used, whereas the size of the labeled data samples varies from 10 to 30,000. Each point on this plot is the result of a single randomly seeded training dataset. For each labeled dataset size, 5 trials were run. The relative error between the CNN and the GAN can be seen in \cref{fig:age-systematic-gan-to-dnn-relative-error}. We see a significant accuracy improvement in every case tested. At 100 labeled examples, the GAN achieves a MAE of 10.6, an accuracy which is not achieved by the CNN until it has 5000 labeled examples available for training. At 100 labeled examples, the GAN has 75\% the error that the CNN does.

The advantage of the SR-GAN drops to near zero as the number of images approaches the number of unlabeled examples being used. There seem to be two likely causes for this. Either, there are enough training images that the network is at capacity (additional images will not further improve the results), or the ratio of labeled to unlabeled images is too small for the generator to be of more benefit to the discriminator. Unfortunately, the number of images available in the IMDB-WIKI dataset make it difficult to pursue a larger number of training examples further.

\subsection{Crowd Counting}
The fourth application we consider is the complex problem of dense crowd counting. Every year, crowds of thousands to millions gather for protests, marathons, pilgrimages, festivals, concerts, and sports events. For each of these events, there is a myriad of reasons to desire to know how many people are present. For those holding the event, both real-time management and future event planning is determined by how many people are present, their current locations, and the intervals at which people are present. For security purposes, evacuations planning and where crowding might be a potential harm to individuals is dependent on the size of the crowds. In journalistic pursuits, the size of a crowd attending an event is often used to measure the significance of the event.

We provide the mean absolute count error (MAE), normalized absolute count error (NAE), and root mean squared count error (RMSE). These are given by the following equations:
\begin{equation}
    \text{MAE} = \frac{1}{N}\sum\limits_{i=1}^{N}\abs{\hat{C}_i - C_i}
\end{equation}
\begin{equation}
\text{NAE} = \frac{1}{N}\sum\limits_{i=1}^{N}\frac{\abs{\hat{C}_i - C_i}}{C_i}
\end{equation}
\begin{equation}
\text{RMSE} = \sqrt{\frac{1}{N}\sum\limits_{i=1}^{N}(\hat{C}_i - C_i)^2}
\end{equation}

\citet{idrees2018composition} showed that a vanilla DenseNet~\citep{huang2017densely} outperformed many application-specific networks for crowd counting. Though \citet{idrees2018composition} then provides an application specific version of DenseNet, we chose to use the vanilla version of DenseNet201 as the discriminator in our experiments. This is done to avoid application specific nuances that distract from the main focus of our work, while still providing a network comparable to the state-of-the-art in terms of accuracy. For the generator, we use the same DCGAN generator architecture as was used in our age and steering angle experiments.

\begin{figure}
    \centering
    \includegraphics[width=0.75\columnwidth]{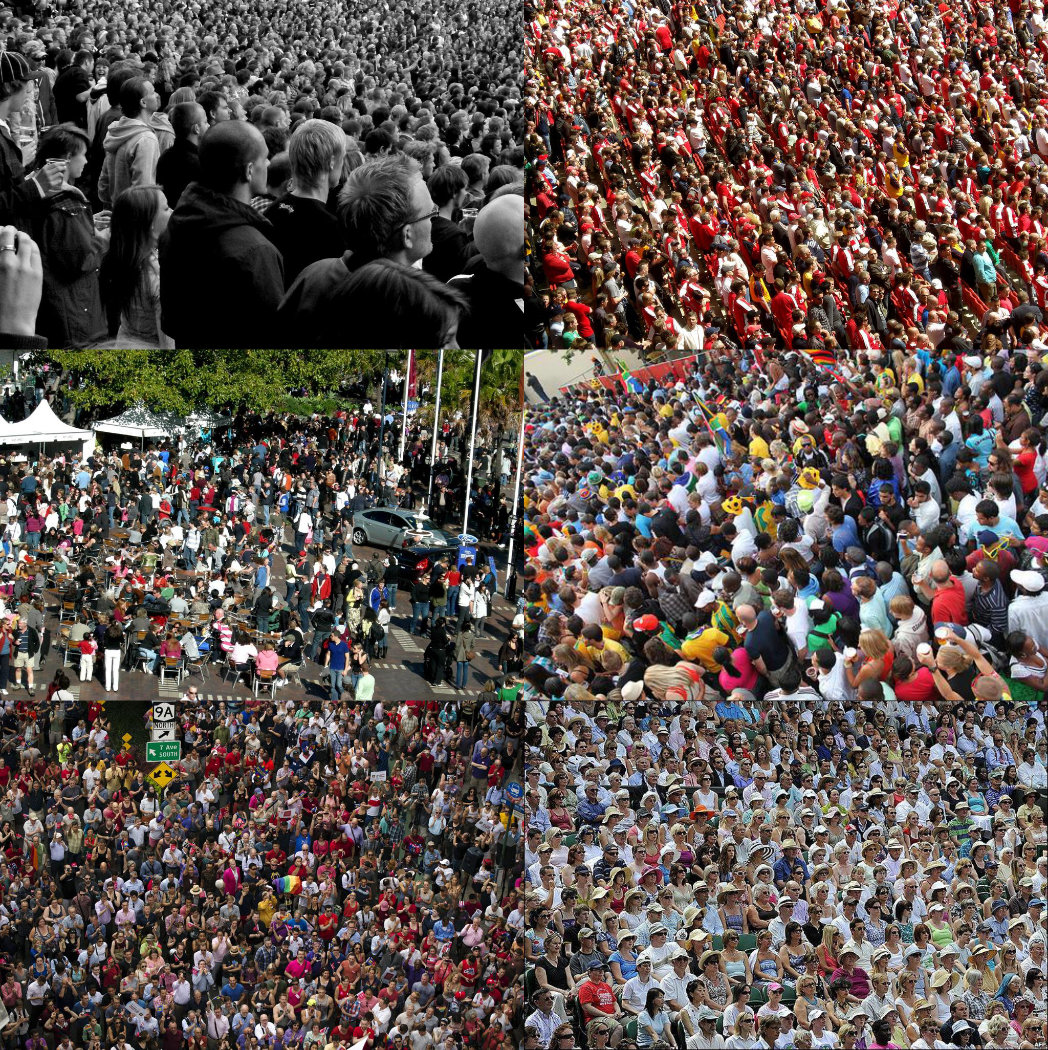}
    \caption{Full image examples from the ShanghaiTech crowd counting dataset.}
    \label{fig:ShanghaiTech Examples}
\end{figure}

\begin{figure}
\centering
\begin{subfigure}{.5\linewidth}
  \centering
  \includegraphics[width=.95\linewidth]{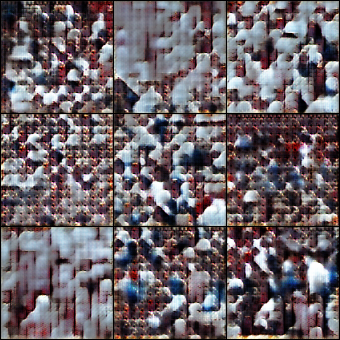}
  \caption{Fake crowd counting images.}
\end{subfigure}%
\begin{subfigure}{.5\linewidth}
  \centering
  \includegraphics[width=.95\linewidth]{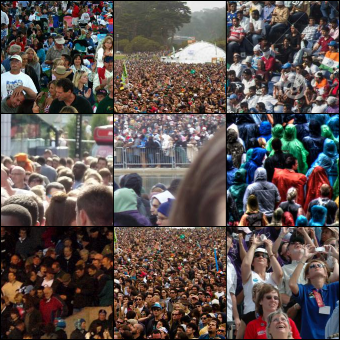}
  \caption{Real crowd counting images.}
\end{subfigure}
\caption{Examples of real and fake images used/generated during training. We note that our approach is not intended to produce realistic looking images, and the fake images are only included for insight.}
\label{fig:Crowd examples}
\end{figure}

The dataset we evaluated our approach on is the ShanghaiTech dataset~\citep{zhang2016single} part A.
The dataset is split into two parts, of which we used Part A in our experiments. Part A contains 482 images, 300 for training and 182 for testing. It contains a total of 241,677 head labelings, with an average of 501.4, a maximum of 3,139, and a minimum of 33.  This part contains a wide range of image sizes, head counts, and perspectives. We used the training and testing images as prescribed by the dataset provider, except we used limited labels for training. A set of example images can be seen in \cref{fig:ShanghaiTech Examples}. During the training process, patches of the images are used. During testing, a sliding window approach is used to calculate the count for each patch with overlapping patches being averaged. A final summing of the average values produces a count for the entire image. Examples of the patches, both real and fake, used/generated during training are shown in \cref{fig:Crowd examples}.

We compare a CNN with the SR-GAN model in our experiments. These results can be seen in \cref{tab:Crowd results}. From the experiments, we can clearly see that the SR-GAN model outperforms the CNN model consistently across various amounts of labeled training images (from 50 to 300), on all three measures. Overall, SR-GAN advantage increases when more training examples are provided. For example, the decreases of MAE of using the SR-GAN versus the CNN are 2.6\%, 3.4\%, 6.0\% to 6.4\% for 50, 100, 200 to 300, respectively. This is slightly contrary to what we might expect, as we would assume the advantage of the SR-GAN to diminish as the number of examples becomes very large. However, the increase is small enough that it may simply be due to chance from dataset selection. The percentage in error decreases are small, but they are comparable to the decrease gained by increasing the size of the labeled dataset. In many cases, the SR-GAN provides an improved over the CNN even with smaller numbers of labeled examples. For example, the SR-GAN with 200 images outperforms the CNN with 300 images. Such improvements are also found in the RMSE for the SR-GAN with 100 and 200 labeled examples compared to the CNN with 200 and 300 examples.
\begin{table}
    \centering
    \begin{tabular}{lrrrr}  
        \toprule
        Method & 50 & 100 & 200 & 300 \\
        \midrule
        CNN MAE & 136.9 & 127.5 & 119.2 & 118.0 \\
        SR-GAN MAE & 133.3 & 123.2 & 112.0 & 110.5 \\
        \midrule
        CNN NAE & 0.342 & 0.354 & 0.359 & 0.357 \\
        SR-GAN NAE & 0.339 & 0.348 & 0.321 & 0.323 \\
        \midrule
        CNN RMSE & 208.5 & 185.1 & 183.2 & 182.3 \\
        SR-GAN RMSE & 205.9 & 178.3 & 178.2 & 169.5 \\
        \bottomrule
    \end{tabular}
    \caption{Crowd counting errors compared various amounts of labeled training examples.}
    \label{tab:Crowd results}
\end{table}

\section{Conclusions}
\label{sec:Conclusion}

Throughout this work, we have presented a means by which to train semi-supervised GANs in a regression situation. The new SR-GAN algorithm was explained in detail. A set of optimization rules which allows for stable, consistent training when using the SR-GAN, including experiments demonstrating the importance of these rules, were given. We performed systematic experiments using the SR-GAN on the real world applications of age estimation,driving steering angle prediction, and crowd counting, all from single images, showing the benefits of SR-GANs over existing approaches. Adding the SR-GAN generator and objectives to a CNN when unlabeled data is available almost always increases the predictive accuracy of the CNN.

We believe this work demonstrates a way in which semi-supervised GANs can be applied generally to a wide range of regression problems with little or no change to the algorithm presented here. This work allows such problems to be solved using deep learning with significantly less labeled training data than was previously required.
\section{Acknowledgments}
\label{sec:Acknowledgments}

This research was initiated under appointments to the U.S. Department of Homeland Security (DHS) Science \& Technology Directorate Office of University Programs, administered by the Oak Ridge Institute for Science and Education (ORISE) through an interagency agreement between the U.S. Department of Energy (DOE) and DHS. ORISE is managed by ORAU under DOE contract number DE-AC05-06OR23100 and DE-SC0014664. All opinions expressed in this paper are the author's and do not necessarily reflect the policies and views of DHS, DOE, or ORAU/ORISE.  The research is also supported by National Science Foundation through Awards PFI \#1827505 and SCC-Planning \#1737533, and Bentley Systems, Incorporated, through a CUNY-Bentley Collaborative Research Agreement (CRA).

\bibliographystyle{model2-names}
\bibliography{bibliography}

\clearpage
\authorbiography[width=3cm, imagewidth=2.6cm,wraplines=10,overhang=0pt]{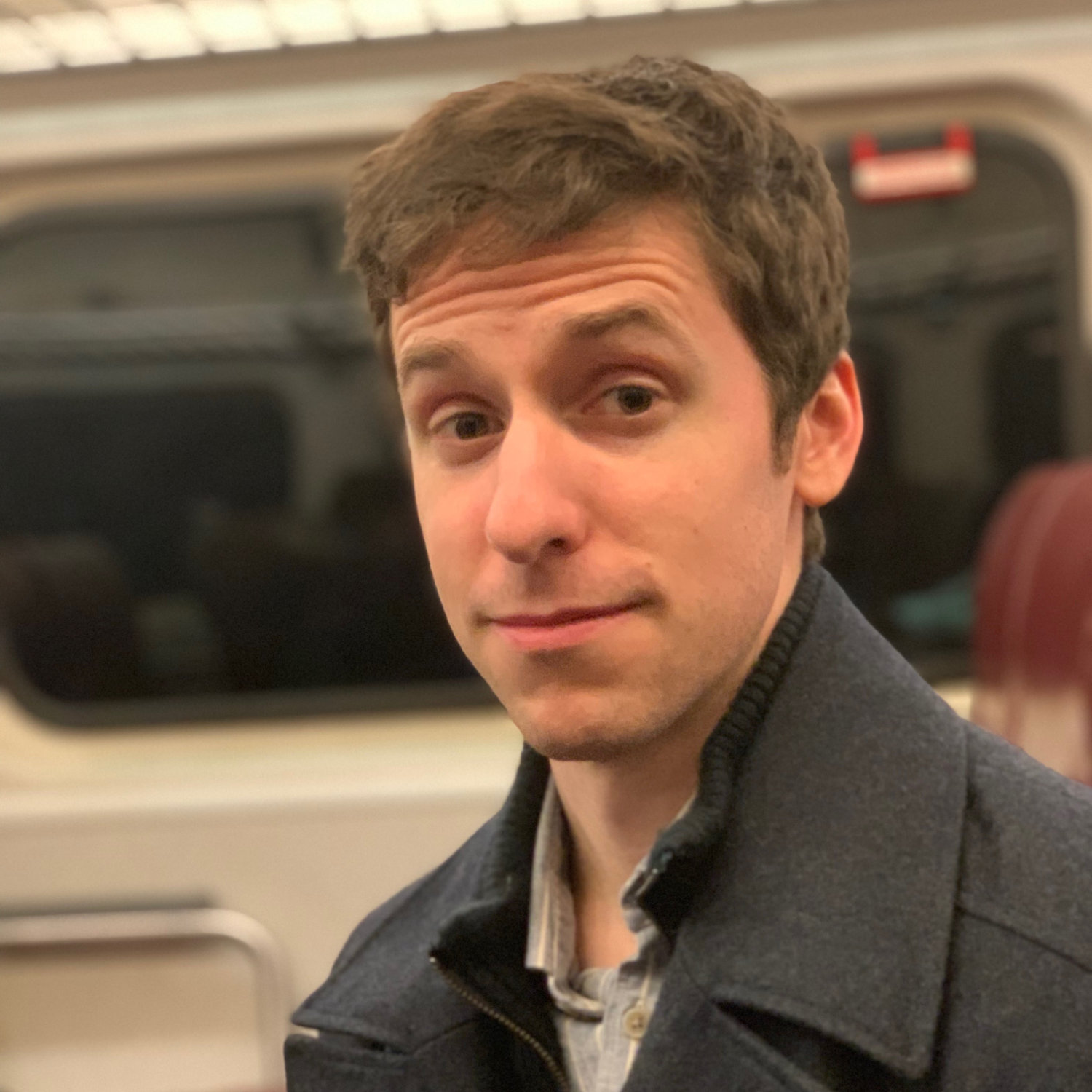}{Greg Olmschenk}{
    Greg Olmschenk is a PhD candidate in Computer Science at the Graduate Center of the City University of New York. His focus is deep neural networks, particularly generative adversarial networks and methods used for computer vision applications.
    \vspace{2\baselineskip}
}%

\authorbiography[width=3cm, imagewidth=2.6cm,wraplines=10,overhang=0pt]{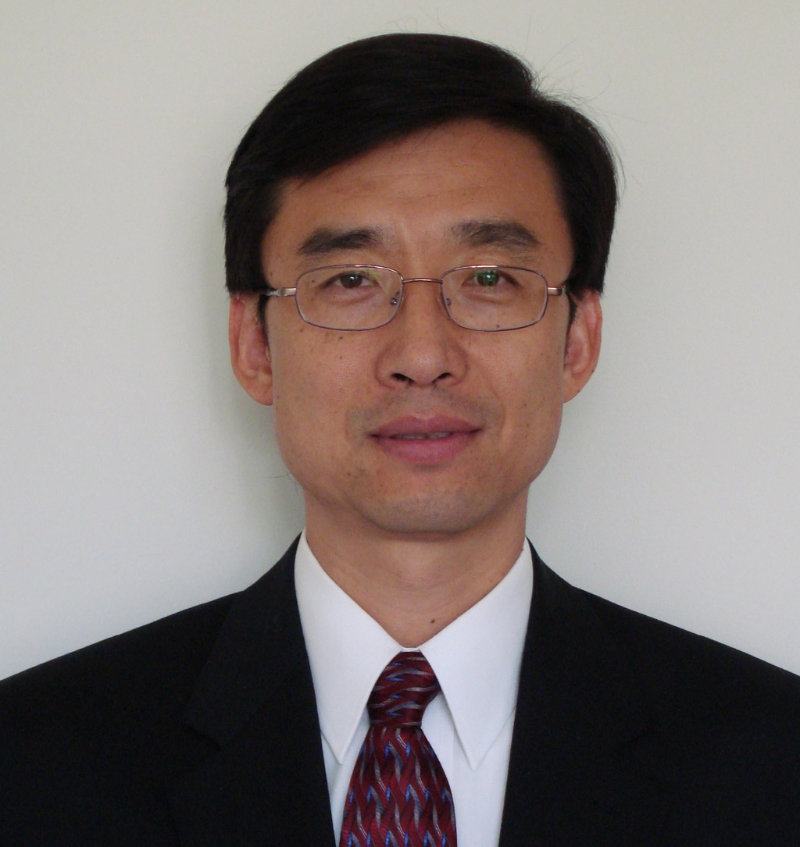}{Zhigang Zhu}{
    Zhigang Zhu received his BE, ME and PhD degrees, all in computer science, from Tsinghua University, Beijing. He is Herbert G. Kayser Chair Professor of Computer Science, at The City College of New York (CCNY) and The CUNY Graduate Center, where he directs the City College Visual Computing Laboratory (CcvcL). His research interests include 3D computer vision, multimodal sensing, virtual/augmented reality, and various applications in assistive technology, robotics, surveillance and transportation. He has published over 150 technical papers in the related fields. He is an Associate Editor of the Machine Vision Applications Journal, Springer, and the IFAC Mechatronics Journal, Elsevier.
}%

\authorbiography[width=3cm, imagewidth=2.6cm,wraplines=10,overhang=0pt]{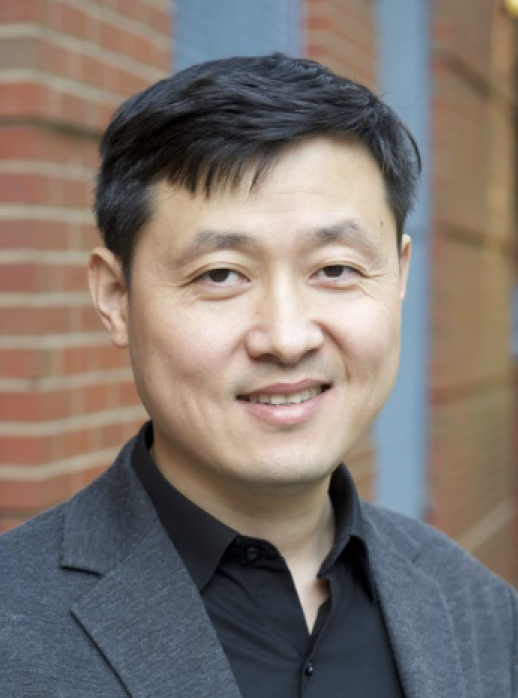}{Hao Tang}{
    Hao Tang is an Associate Professor of Computer Science at The Borough of Manhattan Community College, CUNY. He earned his Ph.D. degree in Computer Science, concentrating in the Computer Vision, at the Graduate Center of CUNY. His research interests are in the fields of 3D computer modeling, HCI, mobile vision and navigation and the applications in surveillance, assistive technology, and education. His research paper was selected as the best paper finalist of International Conference on Multimedia and Expo.
    \vspace{2\baselineskip}
}%

\Authorbiography

\end{document}